\documentclass[11pt, a4paper, logo]{lumia}

\usepackage[numbers,sort&compress]{natbib}
\definecolor{infrecred}{HTML}{D2071E}
\definecolor{abstrabg}{HTML}{FCDDDC}
\colorlet{abstradark}{infrecred}
\definecolor{reportaccent}{HTML}{BB0819}
\colorlet{reporttoc}{reportaccent}
\definecolor{cayenne}{HTML}{941100}
\hypersetup{allcolors=cayenne}
\definecolor{darkblue}{rgb}{0, 0, 0.5}
\definecolor{lightblue}{HTML}{D0CFEA}

\usepackage{graphicx}
\usepackage{subcaption}
\usepackage{multirow}
\usepackage{algorithm}
\usepackage{algorithmic}
\usepackage{titletoc}
\usepackage{bm}

\definecolor{bfgray}{RGB}{247,247,247}
\definecolor{tokenblue}{RGB}{240,247,255}
\definecolor{kvyellow}{RGB}{255,252,238}
\definecolor{oursgreen}{RGB}{240,250,243}

\tcbuselibrary{breakable,listings}
\newtcblisting{benchmarkprompt}[1]{
    breakable,
    listing only,
    title={#1},
    colback=gray!3,
    colframe=black!35,
    colbacktitle=gray!12,
    coltitle=black,
    fonttitle=\small\bfseries,
    boxrule=0.4pt,
    arc=1mm,
    left=2mm,
    right=2mm,
    top=1mm,
    bottom=1mm,
    before skip=8pt,
    after skip=8pt,
    listing options={
        language={},
        basicstyle=\small\ttfamily,
        columns=fullflexible,
        keepspaces=true,
        showstringspaces=false,
        breaklines=true,
        breakatwhitespace=false,
        tabsize=2
    }
}

\usepackage{amsmath,amsfonts,bm}

\def\eqref#1{equation~\ref{#1}}
\def\1{\bm{1}}

\DeclareMathAlphabet{\mathsfit}{\encodingdefault}{\sfdefault}{m}{sl}
\SetMathAlphabet{\mathsfit}{bold}{\encodingdefault}{\sfdefault}{bx}{n}

\setheadertext{R$^2$-Omni}
\setheadertitle{R$^2$-Omni}
\title{\LARGE R$^2$-Omni: Reducing Structural and Precision Redundancy in Omni-LLMs for Long-Form Generation}

\author{%
  \parbox{\textwidth}{\centering
    {\Authfont
      Bolong~Feng$^{1,2}$ \quad
      Pengfei~Zhou$^{1,3,\dagger}$ \quad
      Zhenglin~Wan$^{3}$ \quad
      Wei~Wang$^{4}$ \quad
      Wangbo~Zhao$^{4,\dagger}$\par
    }
    \vspace{0.4em}
    {\Affilfont
      $^{1}$InfRec, Cardinal AI Lab \quad
      $^{2}$Shanghai Jiao Tong University \par
      $^{3}$National University of Singapore \quad
      $^{4}$The Hong Kong University of Science and Technology\par
      $^{\dagger}$Corresponding authors
    }
  }%
}

\begin{document}
% InfRec's class collects the abstract before maketitle, unlike ICLR.
\begin{abstract}
Omni-modal large language models (Omni-LLMs) typically process large numbers of multimodal tokens from audio-visual inputs, incurring substantial computational costs and a large memory footprint for the key-value (KV) cache.
Aggressive token pruning can alleviate this problem while maintaining competitive performance on multiple-choice tasks, which require only a single option as output. However, we observe substantial quality degradation when generating long-form, open-ended responses.
We attribute this gap to query-guided selection, which can retain sufficient evidence for answer selection while discarding details needed for comprehensive generation. This motivates reducing overlapping information across tokens, termed structural redundancy, while preserving diverse content. However, the retained representations may still contain precision redundancy due to unnecessarily high numerical precision in their cached values.
To address these two levels of redundancy in token representations, we introduce a training-free framework named \textbf{R$^2$-Omni}.
First, Structure-Aware Token Consolidation (STC) consolidates redundant representations using spatiotemporal similarity in video and local continuity in audio. Then, Semantic-Sensitivity Guided Precision Allocation (SSPA) allocates value precision to the retained representations according to the estimated reduction in attention-output error under quantized keys, with query attention and audio-visual alignment providing a semantic prior. 
Extensive experiments on long-form generation benchmarks demonstrate an average \textbf{5.0$\times$} reduction in the memory footprint while preserving \textbf{97.98\%} of baseline generation quality and achieving a \textbf{1.48$\times$} end-to-end speedup, highlighting the benefits of jointly reducing structural and precision redundancy.
\end{abstract}
\maketitle

\section{Introduction}
\label{sec:introduction}

Omni-modal large language models (Omni-LLMs)~\citep{qwen25omnitechnicalreport,xu2025qwen3,cui2026minicpm} can jointly process text, video, and audio, enabling powerful audio-visual reasoning and open-ended generation.
These capabilities support diverse applications, including real-time multimodal assistants~\citep{deshmukh2026nemotron,lin2025agent} and complex video understanding tasks~\citep{tang2025videosalmonn2,yao2026omnidcbench}.
However, long audio-visual inputs introduce many multimodal tokens, substantially increasing prefill computation and the memory footprint during autoregressive decoding.
% ⭕️

\begin{figure*}[t]
    \centering
    \includegraphics[width=\textwidth]{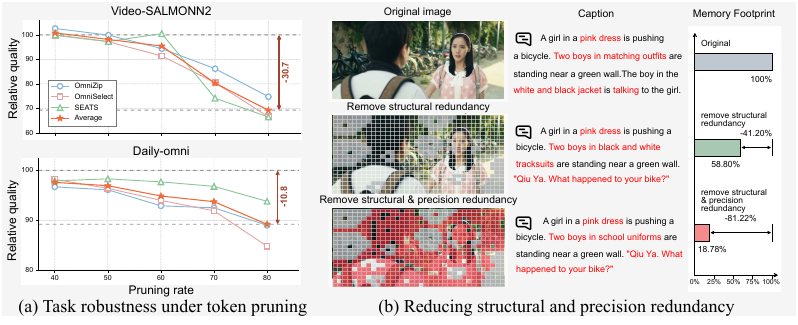}

    \caption{
    \textbf{Motivation of R$^2$-Omni.}
    (a) Aggressive token pruning degrades long-form generation on the Video-SALMONN~2 testset~\citep{tang2025videosalmonn2} more severely than multiple-choice performance on Daily-Omni~\citep{zhou2025daily}. Details are provided in Appendix~\ref{app:fig1}
    (b) After structural redundancy is reduced (\colorbox{gray!20}{gray masks}), we further reduce the precision of cached values (\colorbox{red!20}{red masks}), and the model holds all details. This suggests that precision redundancy remains, enabling further reduction in the memory footprint.
    }

    \label{fig:motivation}
\end{figure*}

Token pruning~\citep{tao2026omnizip,ding2026omnisift,yang2026omniselect,xin2026stage,su2026omnipack} effectively reduces these costs by removing redundant tokens from the multimodal input. These methods can substantially reduce token counts while maintaining strong performance on multiple-choice benchmarks~\citep{fu2025videomme,hong2026worldsense,tao2026lvomnibench,zhou2025daily}, even under aggressive compression. However, this behavior does not extend equally to long-form generation. As shown in Fig.~\ref{fig:motivation}(a), dense-captioning quality degrades much faster than multiple-choice performance as the pruning ratio increases, revealing the limitations of these methods on open-ended tasks that better reflect real-world use.
% ⭕️

We attribute this gap to query-guided token pruning, where methods typically use query attention to assess token importance for the current question and discard tokens with low scores. For multiple-choice tasks, this approach can retain sufficient evidence to identify the correct option while discarding other irrelevant details.
In contrast, long-form generation is an open-ended task and requires broader information coverage, including more events, attributes, and temporal details beyond the evidence needed for a single decision.
Tokens receiving low attention scores from the initial query may still contain unique information needed later in generation, so directly removing them can significantly degrade performance. This suggests that we should target only intrinsic \emph{structural redundancy}, such as spatial similarity, temporal persistence, and local audio continuity, for token reduction while preserving diverse multimodal content for subsequent generation.
% ⭕️

However, even after structural redundancy is reduced, the retained tokens may still be represented at unnecessarily high numerical precision. 
As illustrated in Fig.~\ref{fig:motivation}(b), reducing the precision of their cached values can further decrease the memory footprint while preserving sufficient information to successfully complete the task.
This \emph{precision redundancy} therefore provides an additional opportunity for compression without discarding more tokens~\citep{liu2024kivi,hooper2024kvquant,zandieh2026turboquant,su2026oscar,su2025OTT}. Taken together, these observations reveal two forms of redundancy in token representations: overlapping information across tokens and excess numerical precision in each token's cached value.

To address this problem, we propose \textbf{R$^2$-Omni}, a training-free framework that jointly reduces structural and precision redundancy in the token representations of Omni-LLMs. Specifically, we first introduce Structure-Aware Token Consolidation (STC) to consolidate overlapping representations based on spatiotemporal similarity in video and local continuity in audio, preserving diverse information while effectively shortening the input sequence. Then, to address precision redundancy within the retained representations, we design Semantic-Sensitivity Guided Precision Allocation (SSPA). It uses query attention and audio-visual alignment as a semantic prior, allocating value precision under a memory budget to minimize attention-output error under quantized keys. Through this coordinated design, STC and SSPA exploit the redundancy across input sequence and cached values in Omni-LLMs, while preserving multimodal information essential for long-form generation.
% ⭕️

Extensive experiments on long-form generation benchmarks~\citep{tang2025videosalmonn2,geng2025longvale,yao2026omnidcbench} demonstrate an average \textbf{5.0$\times$} reduction in memory footprint while preserving \textbf{97.98\%} of baseline generation quality and achieving a \textbf{1.48$\times$} end-to-end speedup, highlighting the practical benefits of jointly addressing both forms of redundancy.
% ⭕️

\section{Related Work}

\paragraph{Token Pruning for Omni-LLMs.}
Early token pruning methods mainly target VLMs through attention-guided pruning, informative-token selection, and dynamic sparsification~\citep{chen2024fastV,yang2025visionzip,he2024zipvl}. Another line of work exploits token similarity to merge redundant video representations~\citep{tao2025dycoke,fu2025framefusion,hyun2025sttm,luo2026openmmego}. With the emergence of Omni-LLMs~\citep{qwen25omnitechnicalreport,xu2025qwen3,cui2026minicpm}, recent studies have extended compression to joint audio-video inputs. Omni-specific methods span modality-asymmetric guidance, where one modality guides compression of the other~\citep{tao2026omnizip,ding2026omnisift}, and more joint or modality-adaptive audio-visual selection~\citep{gong2025echoingpixels,yang2026omniselect}. Recently, compression has expanded across the LLM boundary, coordinating pre-LLM structural reduction with layer-wise semantic refinement inside the LLM~\citep{xin2026stage,su2026omnipack}.

\paragraph{KV Cache Quantization.}
KV-cache quantization has progressed from fixed-precision low-bit schemes
toward outlier-aware, token-adaptive, and mixed-precision designs.
Existing methods exploit K/V distribution asymmetry, preserve salient or
outlier components, or adapt precision according to token, layer, or query
importance~\citep{liu2024kivi, hooper2024kvquant, he2024zipcache,
su2025OTT, zhang2026mixkvq}.
More recently, these techniques have been extended to multimodal models,
including ultra-low-bit quantization for VideoLLMs, dynamic multimodal
quantization, and extreme quantization across language, vision-language, and
Omni-LLMs~\citep{tao2025vidkv, fan2026dynamic, su2026oscar}.
In contrast, our method evaluates precision upgrades directly in the
attention-output space and allocates value precision conditioned on both
quantized keys and joint multimodal semantic evidence.

\paragraph{Joint Token--KV Compression.}
Several works jointly explore token and KV compression.
VLM methods reduce prefill computation and KV storage through token sparsification or temporal compression~\citep{he2024zipvl,tao2025dycoke}, combine streaming token reduction with low-bit memory~\citep{chen2026streamingtom}, or couple token pruning with model and sometimes KV quantization~\citep{li2026quota,wang2026qapruner}.
Omni-specific methods also prune audio-video tokens and their corresponding KV entries~\citep{lee2026apack}.
These approaches focus on VLMs, model quantization, or token eviction.
To the best of our knowledge, R$^2$-Omni is the first framework to jointly reduce structural redundancy in audio-video tokens and precision redundancy in the resulting KV cache for efficient Omni-LLMs.

\section{R$^2$-Omni}
\label{sec:method}

\subsection{Overview}
\label{sec:overview}
Omni-LLMs typically comprise vision and audio encoders, modality projectors,
and an LLM backbone.
Given a video, its visual and audio streams $X^v,X^a$ are sampled and
processed by the corresponding encoders to obtain
\begin{equation}
Z^{v}=g_v(X^{v}), \qquad Z^{a}=g_a(X^{a})
\end{equation}
where $Z^v\in\mathbb R^{N_v\times D}$ and
$Z^a\in\mathbb R^{N_a\times D}$ contain $N_v$ video and $N_a$ audio tokens.
The projectors then map these features into the LLM embedding space,
where the tokens are temporally organized and combined with the text prompt.
During prefill, the LLM processes this sequence to construct layer-wise
KV caches, which are repeatedly accessed during autoregressive decoding.
Long audio-visual sequences therefore increase both prefill computation
and the memory footprint of the cache.

\begin{figure*}[t]
  \centering
  \includegraphics[width=\textwidth]{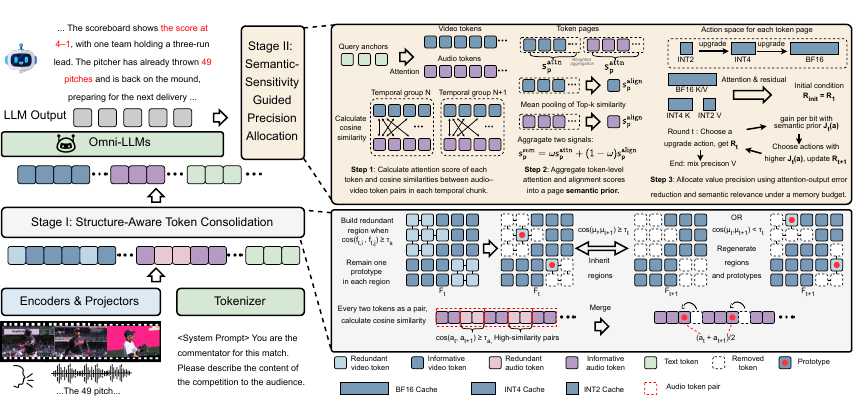}
  \caption{
  \textbf{Overview of R$^2$-Omni.}
  Structure-Aware Token Consolidation(STC) reduces overlapping information across audio-visual tokens before prefill.
  Semantic-Sensitivity Guided Precision Allocation(SSPA) reduces excess precision in the resulting cache by evaluating value
  upgrades through attention-output distortion under quantized keys, guided
  by a multimodal semantic prior.
  }
  \label{fig:method}

\end{figure*}

As shown in Fig.~\ref{fig:method},
R$^2$-Omni applies two training-free stages.
Structure-Aware Token Consolidation (STC) shortens the prefill input by
consolidating overlapping audio-visual content according to intrinsic
similarity rather than initial-query relevance, retaining diverse evidence
for long-form generation.
Semantic-Sensitivity Guided Precision Allocation (SSPA) then assigns
precision to the resulting cache according to attention-output distortion
and a multimodal semantic prior, without evicting additional tokens.
Thus, structural similarity determines consolidation, while output
distortion determines storage precision.

\subsection{Stage \uppercase\expandafter{\romannumeral 1}:
Structure-Aware Token Consolidation}
\label{sec:stage1}
STC exploits spatial and temporal similarity in video and
local continuity in audio. Its consolidation decisions depend on these
input structures, without using query attention as a removal criterion.

\paragraph{Spatial--Temporal Video Consolidation.}
To reduce redundancy within and across frames, we process non-overlapping
adjacent frame pairs $(F_t,F_{t+1})$. Here,
$F_t=\{f_{t,i}\}_{i=1}^{HW}$ and $f_{t,i}\in\mathbb R^D$ is the token
at position $i$ on the $H\times W$ feature grid.
For the first frame $F_t$, we connect spatial neighbors with cosine similarity above
$\tau_s$ and traverse qualifying edges with Union-Find to form connected
components, splitting oversized components into bounded regions.
This confines consolidation to groups formed through local similarity
while retaining isolated tokens. Within each region $R$, we select the
original token closest to its centroid:
\begin{equation}
\bar f_R=\frac{1}{|R|}\sum_{i\in R}f_{t,i},
\qquad
i_R^*=\arg\max_{i\in R}\cos(f_{t,i},\bar f_R).
\label{eq:video_spatial_consolidation}
\end{equation}
We next assess whether these regions persist in the second frame using
the frame means $\mu_t=(HW)^{-1}\sum_i f_{t,i}$.
If $\cos(\mu_t,\mu_{t+1})\geq\tau_t$, we reuse the first frame's region
coverage and remove tokens at the corresponding second-frame positions.
Uncovered positions are retained, and first-frame representatives remain
unchanged. Otherwise, $F_{t+1}$ undergoes independent spatial consolidation.

\paragraph{Local-Continuity Audio Consolidation.}
For audio, structural redundancy arises from locally continuous content.
We therefore divide tokens into local chunks and rank adjacent pairs
$(a_t,a_{t+1})$ satisfying $\cos(a_t,a_{t+1})\geq\tau_a$ by similarity.
We greedily select non-overlapping pairs, replacing each with
$\bar a_t=(a_t+a_{t+1})/2$ at its first position and removing the second.
Chunk boundaries and a cap on the merged-token fraction limit consolidation
of locally continuous content, preserving acoustic details and transitions.
Neighborhood construction, region-size constraints, and position handling
are specified in Appendix~\ref{app:stc_details}.

\subsection{Stage \uppercase\expandafter{\romannumeral 2}:
Semantic-Sensitivity Guided Precision Allocation}
\label{sec:sspa}
After STC, the retained representations may still contain precision redundancy.
SSPA addresses this by evaluating how much attention-output error each
value upgrade removes and allocating precision accordingly.
Following KIVI's K/V asymmetry~\citep{liu2024kivi}, managed historical
keys use channel-wise INT4 and values receive page-wise precision.
We also adopt normalized Hadamard preconditioning~\citep{su2026oscar,ashkboos2024quarot,su2025rotatekv}.

\paragraph{Page Organization and Group-wise Calibration.}
We organize retained tokens into contiguous pages of at most 32 tokens,
splitting pages at modality boundaries.
Tokens within a page share a value precision, making pages common units
for semantic scoring, allocation, and storage.
To estimate their precision needs, we use the final $N_q$ prefill queries
as anchors $Q$. This balances the limited coverage of a single final
query against the cost of using all queries.
These anchors score the retained context before decoding, so calibration
requires no additional generated queries or external calibration data.
For audio--visual alignment, each modality is partitioned into temporal
chunks by relative sequence position, and matching indices define correspondence.
To reduce calibration overhead, we divide decoder layers into groups of four.
The first layer assigns page precision under a depth-dependent budget,
and the remaining layers reuse this assignment while independently
quantizing their own KV.

\paragraph{Multimodal Semantic Prior.}
Query relevance highlights prompt-related content, while audio--visual
alignment captures cross-modal evidence, jointly guiding
precision allocation.
We first measure query relevance and audio--visual alignment at
the token level.
For token $i$, the weighted average of attention received from the
query anchors gives $s_i^{\mathrm{attn}}$.
To capture audio--visual alignment evidence, we compare its normalized
key feature with features from the other modality in the corresponding
temporal chunk. Averaging the largest at most four cosine similarities
gives $s_i^{\mathrm{align}}$.

These token-level signals are then aggregated separately within each page
using a top-$k$ mean and rank-normalized within each modality.
The resulting page scores form the semantic prior
\begin{equation}
s_p^{\mathrm{sem}}
=\omega s_p^{\mathrm{attn}}+(1-\omega)s_p^{\mathrm{align}},
\label{eq:sspa_semantic_prior}
\end{equation}
where $s_p^{\mathrm{attn}}$ and $s_p^{\mathrm{align}}$ are the normalized
page scores and $\omega$ balances query relevance and cross-modal correspondence.
This prior guides upgrades together with output-error reduction and storage cost.
Weighting, aggregation parameters, and fallback rules are given in
Appendix~\ref{app:sspa_details}.

\paragraph{Output-Sensitive Precision Allocation.}
We evaluate precision demand through the effect of a value upgrade on
the attention readout.
At an allocation layer, let $K,V$ be the retained BF16 states and define
$\mathcal A(K)=\operatorname{softmax}(QK^\top/\sqrt d)$, where $d$ is
the attention feature dimension.
A hat denotes reconstruction after quantization.
For each allocation pass, we fix the current layout's INT4 keys $\hat K$
and initialize values in INT2.
This provides a common starting point for evaluating page upgrades.
The initial residual and the readout correction from upgrading page $p$
via action $a=(p,u\!\rightarrow\!v)$ are
\begin{equation}
R_{\mathrm{init}}
=\mathcal A(K)V-\mathcal A(\hat K)\hat V_2,
\qquad
C_a=\mathcal A_p(\hat K)
\left(\hat V_{v,p}-\hat V_{u,p}\right).
\label{eq:sspa_readout_actions}
\end{equation}
Here, $\hat V_2$ is the all-INT2 reconstruction, $\hat V_{b,p}$ is page
$p$ reconstructed at precision $b$, and $\mathcal A_p$ selects its attention
columns. We consider $(u,v)\in\{(2,4),(4,16)\}$, with $\hat V_{16,p}=V_p$.
$R_{\mathrm{init}}$ includes both key-induced attention changes and value
quantization error relative to BF16; an upgrade subtracts $C_a$ from
the current residual.

To account for the effects of previously selected upgrades, allocation
proceeds in rounds $t$ within each refinement pass.
Let $R_t$ denote the residual at the start of round $t$, initialized as
$R_1=R_{\mathrm{init}}$.
The distortion measure $\mathcal L$ emphasizes large anchor errors and
their variation across adjacent anchors (Appendix~\ref{app:sspa_distortion}).
We use this measure to evaluate whether an upgrade's correction reduces
the current residual, defining its benefit and priority as
\begin{equation}
r_t(a)=\mathcal L(R_t)-\mathcal L(R_t-C_a),
\qquad
J_t(a)=\frac{r_t(a)}{\Delta B_a}
\left[1+\beta(2P_a-1)\right].
\label{eq:sspa_upgrade_priority}
\end{equation}
$\Delta B_a$ refers to additional bit cost caused by action $a$, thus $r_t(a)/\Delta B_a$ means the gain from action $a$ per additional bit.
$P_a\in[0,1]$ combines semantic relevance with the within-modality rank
of sensitivity $\mathcal L(C_a)$.
This prioritizes distortion reduction per additional byte, while allowing
semantic relevance and output sensitivity to adjust the ranking of
beneficial upgrades.
Using these priorities, each round greedily selects positive-gain actions
in descending $J_t(a)$, skipping those exceeding the remaining global,
modality, or round budgets. Each page advances by at most one tier.
After applying the selected batch $\mathcal U_t$, we update the residual
as $R_{t+1}=R_t-\sum_{a\in\mathcal U_t}C_a$ and recompute gains for
the next round.

The inner rounds hold $\hat K$ fixed, but the resulting assignment can
affect its reconstruction: value pages are stored in precision-specific
pools, with keys arranged in the same order.
This preserves logical positions while potentially changing the tokens
that share a key quantization block.
Outer passes therefore reconstruct $\hat K$ under the updated layout
and restart allocation from INT2 with recomputed residuals and corrections.
We evaluate each candidate under its own reconstructed KV cache and retain
the lowest-distortion assignment.
Algorithm~\ref{alg:sspa} and cache-layout, budget, and refinement details
are provided in Appendix~\ref{app:sspa_details}.

\section{Experiments}

In this section, we evaluate R$^2$-Omni across three long-form generation
benchmarks and three Omni-LLMs. We introduce the experimental setup in
Sec.~\ref{sec:experimental_setup}, present the main results in
Sec.~\ref{sec:main_results}, analyze efficiency in
Sec.~\ref{sec:efficiency}, and discuss ablation studies in
Sec.~\ref{sec:ablation_studies}.

\subsection{Experimental Setup}
\label{sec:experimental_setup}

\paragraph{Benchmarks.}
We evaluate R$^2$-Omni on three long-form audio-visual generation benchmarks: video-SALMONN-2 testset~\citep{tang2025videosalmonn2}, LongVALE~\citep{geng2025longvale}, and OmniDCBench~\citep{yao2026omnidcbench}.
These benchmarks cover diverse long-form audio-visual generation scenarios,
including detailed video captioning, dense time-aware captioning, and
fine-grained temporal event understanding.

\paragraph{Compared Baselines.}
We compare R$^2$-Omni with representative training-free multimodal compression
methods. \emph{Token Pruning Baselines}: 1) \textbf{OmniZip}~\citep{tao2026omnizip}, 2) \textbf{OmniSelect}~\citep{yang2026omniselect}, and 3) \textbf{SEATS}~\citep{xin2026stage}.
\emph{KV-Cache Quantization Baselines}: 1) \textbf{KIVI}~\citep{liu2024kivi}, 2) \textbf{OTT}~\citep{su2025OTT}, and 3) \textbf{OScaR}~\citep{su2026oscar}.
For a fair comparison, all methods use the same backbone models, multimodal
inputs, and decoding configurations whenever applicable.
We align the compression budget under the same comparable compression ratios.

\paragraph{Implementation Details.}
We evaluate R$^2$-Omni on three representative Omni-LLMs:
Qwen2.5-Omni-3B, Qwen2.5-Omni-7B~\citep{qwen25omnitechnicalreport},
and MiniCPM-o-4.5~\citep{cui2026minicpm}.
All experiments are conducted on NVIDIA H800 80GB GPUs using greedy decoding.
Within each model, all methods use the same inputs, prompts, and generation
settings. 

For Video-SALMONN-2, Qwen uniformly samples 128 frames, while MiniCPM
uses native A/V-aligned sampling with at most 100 frames. LongVALE uses
uniform sampling with at most 100 frames, and OmniDCBench samples at
2 FPS with at most 160 frames. Method configurations and baseline adaptations appear in
Appendix~\ref{app:implementation}.

\begin{table}[!htbp]
\centering
\caption{
\textbf{Comparison with token-pruning and KV-cache quantization methods on
three long-form generation benchmarks.}
Ratio refers to the compression ratio of the final memory footprint relative to the baseline.
\protect\colorbox{bfgray}{\strut Original model},
\protect\colorbox{tokenblue}{\strut Token Pruning},
\protect\colorbox{kvyellow}{\strut KV Quantization}, and
\protect\colorbox{oursgreen}{\strut R$^2$-Omni}
indicate the corresponding method groups.
The best and second-best results among compressed methods are highlighted in
\textbf{bold} and \underline{underline}.
}
\label{tab:main_quality}
\small
\setlength{\tabcolsep}{4.0pt}
\renewcommand{\arraystretch}{1.27}

\resizebox{\textwidth}{!}{%
\begin{tabular}{lccc cc cc c}
\toprule
\multirow{2}{*}{Method}
& \multicolumn{3}{c}{Video-SALMONN~2}
& \multicolumn{2}{c}{LongVALE}
& \multicolumn{2}{c}{OmniDCBench}
& \multirow{2}{*}{\shortstack{Avg. Rel.\\Quality}} \\
\cmidrule(lr){2-4}
\cmidrule(lr){5-6}
\cmidrule(lr){7-8}

& Correct $\uparrow$
& EER $\downarrow$
& Ratio $\uparrow$
& SODA-C $\uparrow$
& Ratio $\uparrow$
& SODA-M $\uparrow$
& Ratio $\uparrow$
& \\
\midrule

% Qwen2.5-Omni-3B
\rowcolor{bfgray}
Qwen2.5-Omni-3B
& 24.45 & 85.66 & 1.00$\times$
& 2.06 & 1.00$\times$
& 5.01 & 1.00$\times$
& 100.0 \\

\rowcolor{tokenblue}
+ OmniSelect
& 15.18 & 95.65 & 5.12$\times$
& 2.20 & 5.08$\times$
& 3.39 & 5.14$\times$
& 81.53 \\

\rowcolor{tokenblue}
+ OmniZip
& 17.49 & 92.79 & 5.12$\times$
& 2.12 & 5.18$\times$
& 4.65 & 5.29$\times$
& 89.89 \\

\rowcolor{tokenblue}
+ SEATS
& 16.76 & 93.04 & 4.88$\times$
& \underline{2.21} & 5.28$\times$
& 4.67 & 5.27$\times$
& 90.28 \\

\rowcolor{kvyellow}
+ KIVI
& 19.42 & 89.64 & 5.29$\times$
& 2.15 & 5.22$\times$
& 3.46 & 5.17$\times$
& 87.10 \\

\rowcolor{kvyellow}
+ OScaR
& 22.32 & \underline{87.82} & 5.11$\times$
& 1.80 & 5.15$\times$
& 3.96 & 5.12$\times$
& 88.81 \\

\rowcolor{kvyellow}
+ OTT
& \underline{22.85} & 87.93 & 5.25$\times$
& 1.71 & 5.21$\times$
& \textbf{5.14} & 5.13$\times$
& \underline{94.12} \\

\rowcolor{oursgreen}
+ \textbf{R$^2$-Omni}
& \textbf{23.54} & \textbf{85.63} & 5.39$\times$
& \textbf{2.23} & 5.34$\times$
& \underline{4.78} & 5.06$\times$
& \textbf{99.99} \\

\midrule

% Qwen2.5-Omni-7B
\rowcolor{bfgray}
Qwen2.5-Omni-7B
& 31.97 & 79.39 & 1.00$\times$
& 1.63 & 1.00$\times$
& 7.32 & 1.00$\times$
& 100.0 \\

\rowcolor{tokenblue}
+ OmniSelect
& 18.74 & 92.82 & 5.10$\times$
& 1.48 & 4.93$\times$
& 4.77 & 4.81$\times$
& 75.03 \\

\rowcolor{tokenblue}
+ OmniZip
& 20.00 & 91.28 & 5.08$\times$
& 1.60 & 4.99$\times$
& 5.42 & 4.88$\times$
& 80.43 \\

\rowcolor{tokenblue}
+ SEATS
& 25.74 & 84.93 & 5.10$\times$
& \underline{1.68} & 4.96$\times$
& 6.44 & 4.90$\times$
& 91.26 \\

\rowcolor{kvyellow}
+ KIVI
& 31.25 & 80.82 & 5.00$\times$
& 1.52 & 5.17$\times$
& 7.07 & 4.97$\times$
& 96.45 \\

\rowcolor{kvyellow}
+ OScaR
& 31.23 & \textbf{78.48} & 5.17$\times$
& 1.61 & 5.12$\times$
& 7.04 & 5.07$\times$
& 98.45 \\

\rowcolor{kvyellow}
+ OTT
& \underline{31.47} & 80.31 & 4.92$\times$
& 1.43 & 5.04$\times$
& \textbf{8.18} & 4.89$\times$
& \underline{99.19} \\

\rowcolor{oursgreen}
+ \textbf{R$^2$-Omni}
& \textbf{31.78} & \underline{79.96} & 5.07$\times$
& \textbf{1.71} & 4.94$\times$
& \underline{7.09} & 4.91$\times$
& \textbf{100.11} \\

\midrule

% MiniCPM-o-4.5
\rowcolor{bfgray}
MiniCPM-o-4.5
& 41.01 & 68.12 & 1.00$\times$
& 2.72 & 1.00$\times$
& 17.14 & 1.00$\times$
& 100.0 \\

\rowcolor{tokenblue}
+ OmniSelect
& 12.87 & 95.02 & 4.92$\times$
& 1.54 & 5.10$\times$
& 3.47 & 4.93$\times$
& 44.98 \\

\rowcolor{tokenblue}
+ OmniZip
& 13.83 & 94.30 & 4.93$\times$
& 1.68 & 5.12$\times$
& 3.39 & 5.21$\times$
& 46.88 \\

\rowcolor{tokenblue}
+ SEATS
& 14.06 & 94.44 & 4.89$\times$
& 1.58 & 5.16$\times$
& 2.87 & 5.37$\times$
& 45.31 \\

\rowcolor{kvyellow}
+ KIVI
& 35.75 & 72.19 & 5.01$\times$
& \underline{2.61} & 5.22$\times$
& 15.01 & 5.26$\times$
& 91.27 \\

\rowcolor{kvyellow}
+ OScaR
& \underline{37.27} & \textbf{70.78} & 4.81$\times$
& \textbf{2.62} & 4.96$\times$
& 15.20 & 5.08$\times$
& \underline{93.03} \\

\rowcolor{kvyellow}
+ OTT
& 36.09 & 71.89 & 4.88$\times$
& 2.53 & 5.15$\times$
& \underline{15.44} & 5.35$\times$
& 91.46 \\

\rowcolor{oursgreen}
+ \textbf{R$^2$-Omni}
& \textbf{37.42} & \underline{71.08} & 5.28$\times$
& 2.33 & 5.08$\times$
& \textbf{17.60} & 4.96$\times$
& \textbf{93.86} \\

\bottomrule
\end{tabular}%
}

\end{table}

\begin{table}[!htbp]
\centering
\caption{
\textbf{Comparison with naively composed token-pruning and KV-cache
quantization baselines.} comp. Ratio refers to the compression ratio of the final memory footprint relative to the baseline.
The best and second-best quality results among completed compressed methods
are highlighted in \textbf{bold} and \underline{underline}.
}
\label{tab:naive_composition}
\small
\setlength{\tabcolsep}{4.0pt}
\renewcommand{\arraystretch}{1.20}

\resizebox{\textwidth}{!}{%
\begin{tabular}{lccc cc cc c}
\toprule
\multirow{2}{*}{Method}
& \multicolumn{3}{c}{Video-SALMONN~2}
& \multicolumn{2}{c}{LongVALE}
& \multicolumn{2}{c}{OmniDCBench}
& \multirow{2}{*}{\shortstack{Avg. Rel.\\Quality}} \\
\cmidrule(lr){2-4}
\cmidrule(lr){5-6}
\cmidrule(lr){7-8}

& Correct $\uparrow$
& EER $\downarrow$
& Ratio $\uparrow$
& SODA-C $\uparrow$
& Ratio $\uparrow$
& SODA-M $\uparrow$
& Ratio $\uparrow$
& \\
\midrule

Qwen2.5-Omni-3B
& 24.45 & 85.66 & 1.00$\times$
& 2.06 & 1.00$\times$
& 5.01 & 1.00$\times$
& 100.0 \\

+ OmniZip + KIVI
& 18.99 & 89.41 & 5.11$\times$
& 1.97 & 5.23$\times$
& 3.79 & 5.07$\times$
& 86.19 \\

+ OmniSelect + OTT
& \underline{22.48} & 88.14 & 5.12$\times$
& 1.92 & 5.37$\times$
& \textbf{5.02} & 5.19$\times$
& \underline{95.63} \\

+ SEATS + OScaR
& 21.44 & \underline{87.49} & 5.32$\times$
& \underline{1.98} & 5.34$\times$
& 2.71 & 5.51$\times$
& 83.95 \\

+ \textbf{R$^2$-Omni}
& \textbf{23.54} & \textbf{85.63} & 5.39$\times$
& \textbf{2.23} & 5.34$\times$
& \underline{4.78} & 5.06$\times$
& \textbf{99.99} \\

\bottomrule
\end{tabular}%
}

\end{table}

\subsection{Main Results on Long-form Generation Benchmarks}
\label{sec:main_results}
We evaluate R$^2$-Omni against token-pruning and KV-cache quantization
baselines on three Omni-LLMs and three long-form generation benchmarks
at a fixed budget. We further compare with naive pruning--quantization compositions and study performance across compression ratios. 

\paragraph{Results across Omni-LLMs.}
Qwen2.5-Omni and MiniCPM-o-4.5 employ different architectures for processing
audio-visual inputs. We preserve their native multimodal input pipelines
and adapt the compression methods to their token and KV-cache interfaces,
enabling comparison under comparable cache budgets. In
Table~\ref{tab:main_quality}, we evaluate our method against existing
pruning and quantization methods on both Qwen2.5-Omni model scales and
MiniCPM-o-4.5. Notably, R$^2$-Omni outperforms all baselines in average
relative quality across the evaluated backbones, demonstrating its ability
to preserve long-form generation quality under compression.

\paragraph{Comparison with Naive Composition.}
Simply combining token pruning and KV-cache quantization can compound
information loss: pruning removes semantic evidence, while quantization
introduces additional errors into the retained representations. In
Table~\ref{tab:naive_composition}, we compare R$^2$-Omni with direct
combinations of existing pruning and quantization methods. Notably,
R$^2$-Omni achieves the highest average relative quality, highlighting the
benefit of a unified design that coordinates token consolidation and
precision allocation to limit their combined distortion. More results across different Omni-LLMs are provided in Appendix~\ref{app:composition_extra}.

\begin{figure}[t]
    \centering

    \begin{minipage}[t]{0.49\linewidth}
        \vspace{0pt}
        \centering
        \includegraphics[width=\linewidth]{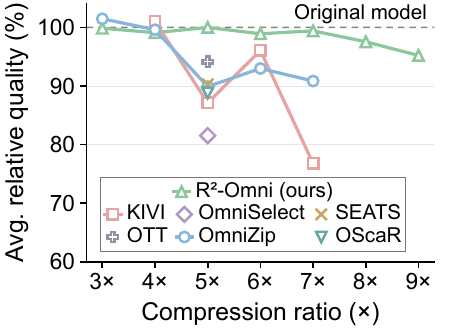}

        \captionof{figure}{
        \textbf{Quality--compression trade-off on Qwen2.5-Omni-3B.}
        Average relative quality across three long-form generation benchmarks. Solid curves connect results at different compression ratios, while unconnected markers indicate single-point results. R$^2$-Omni maintains comparatively stable quality as compression increases. Details are provided in Appendix~\ref{app:rate_results}.
        }
        \label{fig:pareto}
    \end{minipage}
    \hfill
    \begin{minipage}[t]{0.49\linewidth}
        \vspace{0pt}
        \centering
        \includegraphics[width=\linewidth]{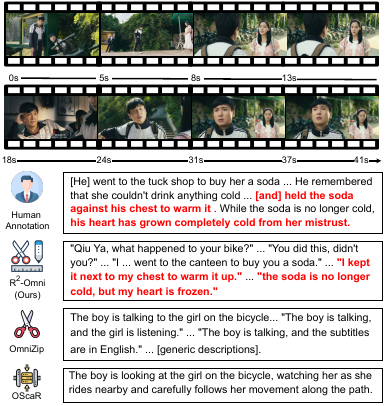}

        \captionof{figure}{
        \textbf{Case study.}
        R$^2$-Omni better preserves the dialogue content
        and its associated multimodal details.
        }
        \label{fig:case}
    \end{minipage}

\end{figure}

\paragraph{Results across Compression Ratios.}
Stronger cache compression places greater pressure on preserving the multimodal evidence needed for long-form generation. Figure~\ref{fig:pareto} shows the results of R$^2$-Omni, KIVI, and OmniZip across different compression ratios on three benchmarks. R$^2$-Omni maintains stable quality as compression increases, while the baselines exhibit more obvious degradation. Notably, R$^2$-Omni retains 95.21\% average relative quality even at $9\times$ compression, demonstrating its ability to achieve higher compression ratio while preserving generation quality.

\paragraph{Case study.}
As shown in Figure~\ref{fig:case}, R$^2$-Omni better preserves speech-dependent and cross-modal semantics, while OmniZip and OScaR mainly retain coarse visual descriptions and lose much of the dialogue information. By preserving complementary visual and audio-text evidence, R$^2$-Omni enables a more coherent understanding of the complete interaction under compression.

\subsection{Efficiency Analysis}
\label{sec:efficiency}

\begin{figure*}[t]
\centering
\includegraphics[width=\textwidth]{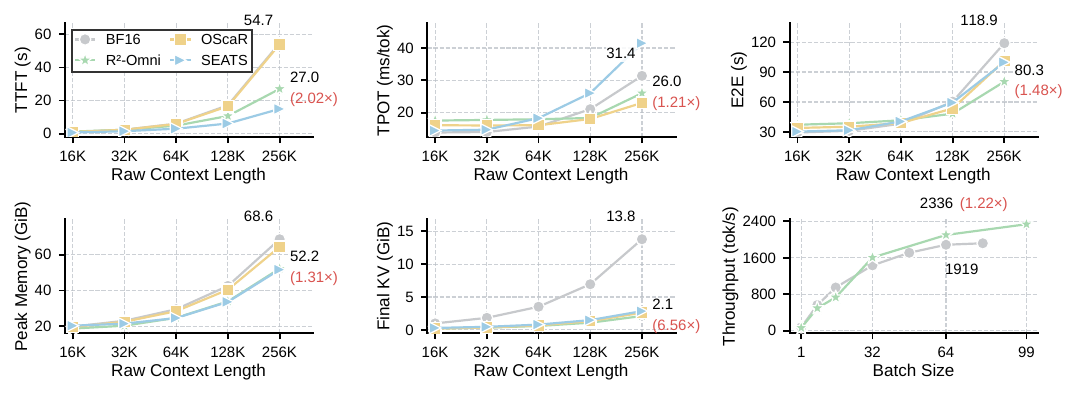}
\caption{
\textbf{End-to-end efficiency of R$^2$-Omni and the BF16 baseline.}
R$^2$-Omni consistently improves inference latency, memory efficiency,
decoding throughput, and maximum batch capacity. Detials are provided in Appendix~\ref{app:efficiency}
}
\label{fig:system_efficiency}

\end{figure*}

We evaluate Qwen2.5-Omni-7B on a single NVIDIA H800 over 16K--256K raw contexts and 2,048 generated tokens.
Figure~\ref{fig:system_efficiency} shows the complementary benefits
of token pruning and KV quantization for prefilling and decoding.
At 256K, R$^2$-Omni achieves the best end-to-end latency, with
$1.48\times$ end-to-end speedup over BF16 and $1.24\times$ over the strongest
competing compression system.
It also provides $1.22\times$ higher peak throughput and
$6.56\times$ final compression of memory footprint.

\begin{table}[!htbp]
\centering
\caption{
\textbf{Component ablation of R$^2$-Omni.} 
Ratio refers to the compression ratio of the final memory footprint relative to the baseline.
Quality is evaluated with 256 frames on Video-SALMONN~2; LongVALE and OmniDCBench follow the main input settings.
}
\label{tab:ablation}
\small
\setlength{\tabcolsep}{2pt}
\renewcommand{\arraystretch}{1.18}

\begin{tabular}{l ccc cc cc ccc}
\toprule
\multirow{2}{*}{Method}
& \multicolumn{3}{c}{Video-SALMONN~2}
& \multicolumn{2}{c}{LongVALE}
& \multicolumn{2}{c}{OmniDCBench}
& \multicolumn{3}{c}{Efficiency (256K)} \\
\cmidrule(lr){2-4}
\cmidrule(lr){5-6}
\cmidrule(lr){7-8}
\cmidrule(lr){9-11}

& Correct $\uparrow$
& EER $\downarrow$
& Ratio
& SODA-C $\uparrow$
& Ratio
& SODA-M $\uparrow$
& Ratio
& TTFT(s) 
& TPOT(ms/tok) 
& E2E(s)  \\
\midrule

\textbf{R$^2$-Omni}
& \textbf{21.02}
& \textbf{88.57}
& 7.07$\times$
& \textbf{2.12}
& 6.79$\times$
& \underline{5.11}
& 6.93$\times$
& \underline{27.03}
& \underline{25.90}
& \underline{40.34} \\

w/o SSPA
& 17.66
& 91.80
& 7.11$\times$
& \underline{2.03}
& 5.13$\times$
& 3.47
& 7.01$\times$
& \textbf{25.50}
& \textbf{21.86}
& \textbf{36.67} \\

w/o STC
& \underline{20.39}
& \underline{90.16}
& 3.97$\times$
& 1.77
& 3.95$\times$
& \textbf{5.35}
& 3.95$\times$
& 58.54
& 39.61
& 78.86 \\

\bottomrule
\end{tabular}%
\end{table}

\subsection{Ablation Studies}
\label{sec:ablation_studies}
In this section, we conduct ablation studies to examine the contributions of STC and SSPA,
the role of semantic signals in precision allocation, and the
compression--quality trade-off induced by STC similarity thresholds.

\paragraph{Contributions of STC and SSPA.}
STC and SSPA address redundancy in the token sequence and the precision
of the retained representations, respectively. In Table~\ref{tab:ablation},
we compare the full method with variants that retain only one stage.
Notably, R$^2$-Omni preserves better generation quality than STC-only while
achieving stronger cache compression than SSPA-only. These results
highlight the complementary roles of both stages and support their joint
design for efficient long-form generation.

\begin{table}[!htbp]
\centering
\caption{
\textbf{Ablation of SSPA allocation signals on Qwen2.5-Omni-3B.}
Ratio refers to the compression ratio of the final memory footprint relative to the baseline.
All variants use the same STC output and configured precision budget.
}
\label{tab:sspa_signal_ablation}
\small
\setlength{\tabcolsep}{7pt}
\renewcommand{\arraystretch}{1.18}
\begin{tabular}{@{}lccccccc@{}}
\toprule
\multirow{2}{*}{Method}
& \multicolumn{3}{c}{Video-SALMONN~2}
& \multicolumn{2}{c}{LongVALE}
& \multicolumn{2}{c}{OmniDCBench} \\
\cmidrule(lr){2-4}
\cmidrule(lr){5-6}
\cmidrule(lr){7-8}
& Correct $\uparrow$
& EER $\downarrow$
& \shortstack{Ratio}
& SODA-C $\uparrow$
& \shortstack{Ratio}
& SODA-M $\uparrow$
& \shortstack{Ratio} \\
\midrule

R$^2$-Omni
& \underline{23.54} & \textbf{85.63} & 5.39$\times$
& \textbf{2.23} & 5.34$\times$
& \textbf{4.78} & 5.06$\times$\\ 

w/o A--V Alignment
& \textbf{23.57} & \underline{86.81} & 5.36$\times$
& \underline{2.12} & 5.34$\times$
& 4.55 & 5.00$\times$\\ 

w/o Semantic Prior
& 23.48 & 87.99 & 5.34$\times$
& 2.01 & 5.33$\times$
& \underline{4.61} & 5.04$\times$\\ 

\bottomrule
\end{tabular}
\end{table}

\begin{table}[!htbp]
\centering
\caption{
\textbf{Token retention and generation quality across STC similarity
thresholds on Qwen2.5-Omni-3B.}
Token Ret. refers to Token retention ratio after STC.
}
\label{tab:stc_threshold_ablation}

\small
\setlength{\tabcolsep}{4.5pt}
\renewcommand{\arraystretch}{1.18}

\resizebox{\linewidth}{!}{%
\begin{tabular}{cc ccc cc cc}
\toprule
\multicolumn{2}{c}{Threshold}
& \multicolumn{3}{c}{Video-SALMONN~2}
& \multicolumn{2}{c}{LongVALE}
& \multicolumn{2}{c}{OmniDCBench} \\
\cmidrule(lr){1-2}
\cmidrule(lr){3-5}
\cmidrule(lr){6-7}
\cmidrule(lr){8-9}

$\tau_s$ & $\tau_a$
& Token Ret. (\%)
& Correct $\uparrow$
& EER $\downarrow$
& Token Ret. (\%)
& SODA-C $\uparrow$
& Token Ret. (\%)
& SODA-M $\uparrow$ \\
\midrule

0.7 & 0.8
& 56.14 & 23.54 & 85.63
& 59.73 & 2.23
& 61.84 & 4.78 \\

0.5 & 0.6
& 18.49 & 18.32 & 90.45
& 23.50 & 2.28
& 22.77 & 4.26 \\

0.3 & 0.4
& 6.97 & 16.00 & 92.56
& 12.24 & 2.00
& 10.32 & 3.18 \\

\bottomrule
\end{tabular}%
}
\end{table}

\paragraph{SSPA Allocation Signals.}
The semantic prior guides precision allocation using query attention and
cross-modal relevance. In Table~\ref{tab:sspa_signal_ablation}, we evaluate
its contribution by removing A--V alignment or the entire semantic prior
while preserving the STC output and precision budget. Incorporating both
signals generally improves generation quality, highlighting the benefit
of using semantic information to refine output-sensitive allocation and
preserve relevant content during quantization.

\paragraph{Compression--Quality Trade-off in STC.}
Table~\ref{tab:stc_threshold_ablation} examines how the spatial and audio
similarity thresholds affect token retention and generation quality, with
all other settings fixed.
Lower thresholds admit less similar tokens for consolidation and consistently
reduce token retention.
Quality varies across benchmarks at intermediate thresholds, but the lowest thresholds consistently degrade quality, suggesting that aggressive consolidation may merge distinct content. This motivates using STC to reduce overlapping representations and SSPA to further compress the retained context.

\section{Conclusion}
We presented R$^2$-Omni, a training-free framework that jointly reduces structural and precision redundancy for efficient long-form generation in Omni-LLMs. STC consolidates overlapping audio-visual representations while preserving diverse content, and SSPA allocates value precision according to attention-output sensitivity under quantized keys, guided by a multimodal semantic prior. Across three Omni-LLMs and three long-form generation benchmarks, R$^2$-Omni achieves an average $5.0\times$ KV-cache compression while retaining $97.98\%$ of baseline generation quality. It also delivers a $1.48\times$ end-to-end speedup in our long-context efficiency evaluation. These results highlight the benefits of jointly reducing overlapping representations and excess numerical precision while preserving the multimodal information needed for comprehensive generation.
\paragraph{Limitations and Future Work.}
The achievable compression of R$^2$-Omni depends in part on its precision allocation strategy, which shapes the balance between memory efficiency and generation quality. This balance may vary across models and application settings, leaving room for further refinement of the framework. Future work will extend R$^2$-Omni to a broader range of model architectures and explore its application in agentic systems with more diverse generation requirements.

\subsection*{AI use statement}
We used AI tools to assist with writing and language refinement,
literature retrieval and discovery, and code revision and debugging
during research execution.
The authors take full responsibility for the final content of this work.

\bibliography{iclr2027_conference}

% ============================================================
% Supplementary Material
% ============================================================

\clearpage
\appendix

\startcontents[supplementary]

\begin{center}
    {\LARGE \textbf{Supplementary Material}}
\end{center}

\vspace{1em}

\vspace{0.5em}

\printcontents[supplementary]{}{1}{
    \setcounter{tocdepth}{2}
}

\clearpage

\section{Complete Experimental Results}
\label{app:full_results}

\subsection{Complete Results of token pruning methods across tasks}
\label{app:fig1}
We evaluate all token pruning methods on Qwen2.5-Omni-3B under the same model and evaluation settings. As shown in Table~\ref{tab:token_compression_quality}, moderate token pruning largely preserves model quality, whereas aggressive pruning causes substantial degradation, particularly on dense-captioning tasks. At pruning ratios of 70--80\%, the Correct rate on Video-SALMONN~2 drops markedly across all methods, while the degradation on DailyOmni is comparatively smaller. This indicates that multiple-choice tasks can often be solved using a limited subset of salient evidence, whereas dense captioning requires retaining substantially more fine-grained visual information. These results support our motivation that aggressive token pruning may remove details that are unnecessary for answering a specific question but remain important for comprehensive long-form generation.

\begin{table}[t]
\centering
\caption{
Token pruning quality on Video-SALMONN~2 and DailyOmni.
Video-SALMONN~2 reports Correct rate, while DailyOmni reports multiple-choice accuracy.
Relative quality is normalized to the corresponding BF16 baseline.
}
\label{tab:token_compression_quality}
\small
\setlength{\tabcolsep}{4.5pt}

\begin{tabular}{lccccc}
\toprule
& & \multicolumn{2}{c}{Video-SALMONN~2}
& \multicolumn{2}{c}{DailyOmni} \\
\cmidrule(lr){3-4} \cmidrule(lr){5-6}
Method & Pruning
& Correct $\uparrow$ & Rel. Quality $\uparrow$
& Accuracy $\uparrow$ & Rel. Quality $\uparrow$ \\
\midrule

\rowcolor{black!10}
BF16 & 0\% & 24.45 & 100.00 & 55.56 & 100.00 \\

\midrule

\multirow{5}{*}{OmniZip}
& 40\% & 25.09 & 102.60 & 53.72 & 96.69 \\
& 50\% & 24.43 &  99.90 & 53.38 & 96.09 \\
& 60\% & 23.09 &  94.41 & 51.63 & 92.93 \\
& 70\% & 21.07 &  86.18 & 51.38 & 92.48 \\
& 80\% & 18.28 &  74.76 & 49.46 & 89.02 \\

\midrule

\multirow{5}{*}{OmniSelect}
& 40\% & 24.40 & 99.78 & 54.55 & 98.20 \\
& 50\% & 23.77 & 97.21 & 53.55 & 96.39 \\
& 60\% & 22.34 & 91.37 & 52.21 & 93.98 \\
& 70\% & 19.68 & 80.47 & 51.04 & 91.88 \\
& 80\% & 16.29 & 66.60 & 47.12 & 84.81 \\

\midrule

\multirow{5}{*}{SEATS}
& 40\% & 24.45 & 100.00 & 54.39 & 97.89 \\
& 50\% & 23.80 &  97.33 & 54.64 & 98.35 \\
& 60\% & 24.57 & 100.47 & 54.30 & 97.74 \\
& 70\% & 18.14 &  74.17 & 53.80 & 96.84 \\
& 80\% & 16.26 &  66.48 & 52.13 & 93.83 \\

\bottomrule
\end{tabular}
\end{table}

\subsection{Complete Results of Efficiency Experiment}
\label{app:efficiency}
\begin{table}[t]
\centering
\setlength{\tabcolsep}{5.5pt}
\renewcommand{\arraystretch}{1.08}
\caption{\textbf{End-to-end efficiency across context lengths.}
TTFT denotes time-to-first-token, TPOT denotes time-per-output-token, and
E2E denotes the total latency for generating 2,048 tokens.
Peak is the peak allocated GPU memory, and Final is the physical memory footprint
after generation.
Context lengths refer to raw multimodal sequences before token consolidation.
Lower is better for all metrics.}
\label{tab:system_efficiency}
\resizebox{\linewidth}{!}{
\begin{tabular}{lcrrrrr}
\toprule
\textbf{Method}
& \textbf{Context}
& \textbf{TTFT (s)}
& \textbf{TPOT (ms/tok)}
& \textbf{E2E (s)}
& \textbf{Peak (GiB)}
& \textbf{Final (GiB)} \\
\midrule

BF16
& 16K
& 0.964
& \textbf{13.871}
& \textbf{29.357}
& 19.920
& 0.978 \\
R$^2$-Omni
& 16K
& 1.223
& 17.563
& 37.184
& \textbf{18.790}
& 0.209 \\
OScaR
& 16K
& 0.950
& 16.124
& 33.963
& 19.654
& \textbf{0.191} \\
SEATS
& 16K
& \textbf{0.645}
& 14.482
& 30.288
& 20.387
& 0.288 \\

\midrule

BF16
& 32K
& 2.364
& \textbf{14.037}
& \textbf{31.090}
& 23.134
& 1.824 \\
R$^2$-Omni
& 32K
& 2.305
& 17.757
& 38.653
& \textbf{20.402}
& \textbf{0.334} \\
OScaR
& 32K
& 2.312
& 16.046
& 35.163
& 22.602
& 0.350 \\
SEATS
& 32K
& \textbf{1.442}
& 14.742
& 31.612
& 21.551
& 0.457 \\

\midrule

BF16
& 64K
& 6.013
& \textbf{15.721}
& \textbf{38.196}
& 29.560
& 3.516 \\
R$^2$-Omni
& 64K
& 4.816
& 17.972
& 41.605
& 24.792
& \textbf{0.590} \\
OScaR
& 64K
& 5.829
& 16.114
& 38.814
& 28.502
& 0.668 \\
SEATS
& 64K
& \textbf{2.966}
& 18.315
& 40.461
& \textbf{24.669}
& 0.795 \\

\midrule

BF16
& 128K
& 17.145
& 21.088
& 60.312
& 42.561
& 6.938 \\
R$^2$-Omni
& 128K
& 10.653
& 18.364
& \textbf{48.283}
& 33.922
& \textbf{1.086} \\
OScaR
& 128K
& 16.544
& \textbf{18.059}
& 53.516
& 40.441
& 1.319 \\
SEATS
& 128K
& \textbf{6.100}
& 26.043
& 59.410
& \textbf{33.663}
& 1.479 \\

\midrule

BF16
& 256K
& 54.701
& 31.375
& 118.910
& 68.561
& 13.782 \\
R$^2$-Omni
& 256K
& 27.030
& 26.014
& \textbf{80.347}
& 52.189
& \textbf{2.102} \\
OScaR
& 256K
& 54.023
& \textbf{23.025}
& 101.139
& 64.317
& 2.612 \\
SEATS
& 256K
& \textbf{14.923}
& 41.465
& 99.782
& \textbf{51.652}
& 2.848 \\

\bottomrule
\end{tabular}
}
\end{table}

We evaluate Qwen2.5-Omni-7B on a single NVIDIA H800 GPU using greedy
decoding and raw context lengths from 16K to 256K.
All methods use FlashAttention-2 for prefilling, and each result is the median
of three measured runs after one warm-up run.
As shown in Table~\ref{tab:system_efficiency}, compression introduces
additional selection, quantization, and cache-management overhead, which can
dominate at short context lengths.
Consequently, BF16 remains competitive for short inputs.

The benefit becomes more pronounced as the context grows.
SEATS achieves the lowest TTFT through aggressive token reduction, whereas
OScaR obtains the lowest TPOT at 256K through KV quantization.
By jointly reducing token and precision redundancy, R$^2$-Omni achieves the
best end-to-end latency at both 128K and 256K while achieving the smallest
final memory footprint.
At 256K, it provides a \textbf{1.48$\times$} E2E speedup,
uses \textbf{1.31$\times$} less peak memory, and reduces the final memory footprint
by \textbf{6.56$\times$} over BF16.
It is also \textbf{1.24$\times$} and \textbf{1.26$\times$} faster end-to-end
than SEATS and OScaR, respectively.
These results show that R$^2$-Omni does not optimize a single isolated stage;
instead, it provides a better overall balance between long-context latency and
memory efficiency.

\subsection{Complete Results under Different Compression Ratios}
\label{app:rate_results}

Table~\ref{tab:compression_sweep} presents detailed quality results for
R$^2$-Omni, OmniZip, and KIVI across different compression ratios on
Qwen2.5-Omni-3B. R$^2$-Omni maintains comparatively stable overall quality
as compression increases, while the baselines exhibit more pronounced
degradation under tighter memory budgets. These results demonstrate
the ability of R$^2$-Omni to accommodate stronger memory constraints
while preserving long-form generation quality.

\begin{table}[!htbp]
\centering
\caption{
\textbf{Quality across compression ratios on Qwen2.5-Omni-3B.}
For R$^2$-Omni, each operating point adjusts only the STC thresholds
and SSPA value-precision budget.
}
\label{tab:compression_sweep}
\label{tab:app_baseline_sweep}
\small
\setlength{\tabcolsep}{7pt}
\renewcommand{\arraystretch}{1.12}

\resizebox{0.85\textwidth}{!}{%
\begin{tabular}{@{}lccccc@{}}
\toprule
\multirow{2}{*}{Method}
& \multirow{2}{*}{Compression}
& \multicolumn{2}{c}{Video-SALMONN~2}
& LongVALE
& OmniDCBench \\
\cmidrule(lr){3-4}
\cmidrule(lr){5-5}
\cmidrule(lr){6-6}
& & Correct $\uparrow$
& EER $\downarrow$
& SODA-C $\uparrow$
& SODA-M $\uparrow$ \\
\midrule

\multirow{6}{*}{R$^2$-Omni}
& 3$\times$ & 25.75 & 84.28 & 2.09 & 4.56 \\
& 4$\times$ & 24.02 & 86.05 & 2.13 & 4.77 \\
& 6$\times$ & 22.58 & 87.33 & 2.25 & 4.81 \\
& 7$\times$ & 22.86 & 86.30 & 2.12 & 5.11 \\
& 8$\times$ & 22.18 & 87.02 & 2.11 & 4.94 \\
& 9$\times$ & 21.75 & 87.39 & 2.02 & 4.80 \\

\midrule
\multirow{4}{*}{OmniZip}
& 3$\times$ & 21.39 & 88.08 & 2.50 & 5.00 \\
& 4$\times$ & 20.02 & 89.42 & 2.52 & 4.93 \\
& 6$\times$ & 16.98 & 93.45 & 2.44 & 4.63 \\
& 7$\times$ & 16.01 & 94.35 & 2.49 & 4.32 \\

\midrule
\multirow{3}{*}{KIVI}
& 4$\times$ & 24.51 & 86.09 & 2.04 & 5.28 \\
& 6$\times$ & 22.27 & 87.39 & 2.06 & 4.77 \\
& 7$\times$ & 18.71 & 90.17 & 1.74 & 2.56 \\

\bottomrule
\end{tabular}%
}
\end{table}

\subsection{Results of naive combinations across Omni-LLMs.}
\label{app:composition_extra}

Table~\ref{tab:app_compositions} extends the composition comparison to
Qwen2.5-Omni-7B and MiniCPM-o-4.5. On MiniCPM-o-4.5, SEATS + OScaR
achieves the best results on all four quality metrics among the three
compositions. This reverses its ranking on Qwen2.5-Omni-7B, showing that
the effectiveness of a direct composition depends on the backbone.

\begin{table}[!htbp]
\centering
\caption{\textbf{Direct pruning--quantization compositions on Qwen2.5-Omni-7B and MiniCPM-o-4.5.}}
\label{tab:app_compositions}
\small
\setlength{\tabcolsep}{3pt}
\renewcommand{\arraystretch}{1.12}
\begin{tabular}{@{}lccccccc@{}}
\toprule
\multirow{2}{*}{Method}
& \multicolumn{3}{c}{Video-SALMONN~2}
& \multicolumn{2}{c}{LongVALE} & \multicolumn{2}{c}{OmniDCBench} \\
\cmidrule(lr){2-4}\cmidrule(lr){5-6}\cmidrule(lr){7-8}
& Correct $\uparrow$ & EER $\downarrow$ & KV Comp.
& SODA-C $\uparrow$ & KV Comp. & SODA-M $\uparrow$ & KV Comp. \\
\midrule
\multicolumn{8}{l}{Qwen2.5-Omni-7B} \\
OmniZip + KIVI & 29.97 & 78.57 & 5.07$\times$ & 1.44 & 5.34$\times$ & 6.33 & 4.89$\times$ \\
OmniSelect + OTT & 30.99 & 79.07 & 5.00$\times$ & 1.40 & 5.35$\times$ & 6.84 & 4.81$\times$ \\
SEATS + OScaR & 15.12 & 93.92 & 5.27$\times$ & 1.15 & 5.30$\times$ & 3.57 & 5.22$\times$ \\
\midrule
\multicolumn{8}{l}{MiniCPM-o-4.5} \\
OmniZip + KIVI & 28.97 & 79.26 & 4.90$\times$ & 1.84 & 5.41$\times$ & 16.16 & 5.06$\times$ \\
OmniSelect + OTT & 28.76 & 79.26 & 4.79$\times$ & 1.97 & 5.01$\times$ & 16.63 & 4.93$\times$ \\
SEATS + OScaR & 34.80 & 73.53 & 4.91$\times$ & 2.12 & 4.94$\times$ & 16.87 & 4.93$\times$ \\
\bottomrule
\end{tabular}
\end{table}

\section{Implementation Details}
\label{app:implementation}

\subsection{R$^2$-Omni Implementation}
\label{app:r2omni_implementation}

\subsubsection{Structure-Aware Token Consolidation}
\label{app:stc_details}

STC operates on the projected multimodal embeddings before the first
language-decoder layer.
The representative selection and audio merging rules are described in
Sec.~\ref{sec:stage1}; below we specify region construction, boundary cases,
and position handling.

\paragraph{Spatial region construction.}
We recover the post-vision-merge token grid and use fixed 8-neighbor
connectivity.
Each undirected edge is enumerated once using the offsets
$(0,1)$, $(1,-1)$, $(1,0)$, and $(1,1)$, with out-of-bounds neighbors skipped.
Cosine similarities are computed in a batch, and edges with similarity
strictly greater than $\tau_s$ are processed by Union-Find.
The implementation uses path halving in \textsc{Find}; unions attach one
root to the other without rank- or size-based balancing.

After all qualifying edges have been processed, tokens are grouped by
their final roots.
Each component is sorted by flattened grid index and split into groups
of at most 128 tokens.
Groups containing a single token are retained unchanged; each larger
group retains one original representative according to
Eq.~\ref{eq:video_spatial_consolidation}.
Connectivity is defined before this size-based split: it does not require
pairwise similarity among all members, and the resulting bounded groups
need not remain connected.
Algorithm~\ref{alg:stc_union_find} summarizes this procedure.
In the CUDA implementation, edge similarities are computed on the GPU
and the threshold decisions are transferred to the CPU for Union-Find.

\begin{algorithm}[t]
\caption{Union-Find Spatial Consolidation in STC}
\label{alg:stc_union_find}
\small
\begin{algorithmic}[1]
\REQUIRE Frame features $F=\{f_i\}_{i=1}^{HW}$; token grid;
threshold $\tau_s$; maximum group size $L=128$
\ENSURE Retained indices $\mathcal I$; region groups $\mathcal R$

\STATE \textbf{Helper routine} \textsc{Find}$(i)$:
\WHILE{$\mathrm{parent}[i]\neq i$}
  \STATE $\mathrm{parent}[i]\leftarrow
  \mathrm{parent}[\mathrm{parent}[i]]$
  \STATE $i\leftarrow\mathrm{parent}[i]$
\ENDWHILE
\STATE \textbf{return} $i$
\STATE \textbf{End helper routine}

\STATE \textbf{Main procedure:} initialize $\mathrm{parent}[i]\leftarrow i$
\STATE Enumerate the non-duplicate 8-neighbor edges $\mathcal E$
\STATE Compute $s_{ij}=\cos(f_i,f_j)$ for all $(i,j)\in\mathcal E$
\FORALL{$(i,j)\in\mathcal E$ with $s_{ij}>\tau_s$}
  \STATE $r_i\leftarrow\textsc{Find}(i)$;
  $r_j\leftarrow\textsc{Find}(j)$
  \IF{$r_i\neq r_j$}
    \STATE $\mathrm{parent}[r_j]\leftarrow r_i$
  \ENDIF
\ENDFOR

\STATE Group tokens by $\textsc{Find}(i)$ to obtain components $\mathcal C$
\STATE $\mathcal I\leftarrow\varnothing$;
$\mathcal R\leftarrow\varnothing$
\FORALL{$C\in\mathcal C$}
  \STATE Sort $C$ by grid index and split it into groups of size at most $L$
  \FORALL{resulting groups $R$}
    \IF{$|R|=1$}
      \STATE $\mathcal I\leftarrow\mathcal I\cup R$
    \ELSE
      \STATE Select $i_R^*$ using
      Eq.~\ref{eq:video_spatial_consolidation}
      \STATE $\mathcal I\leftarrow\mathcal I\cup\{i_R^*\}$
      \STATE $\mathcal R\leftarrow\mathcal R\cup\{R\}$
    \ENDIF
  \ENDFOR
\ENDFOR
\STATE \textbf{return} $\mathcal I$ in original order and $\mathcal R$
\end{algorithmic}
\end{algorithm}

\paragraph{Temporal region reuse.}
Video frames are processed in non-overlapping adjacent pairs.
When a pair passes the frame-similarity test in Sec.~\ref{sec:stage1},
the second frame drops positions covered by the first frame's region
groups $\mathcal R$ and retains all remaining positions.
No second-frame representative is selected for these regions, and their
features are not averaged into the first-frame representatives.
If the frame-similarity test fails, the second frame is consolidated
independently. An unpaired final frame undergoes spatial consolidation only.

\paragraph{Local audio merging.}
Audio chunks contain at most 50 tokens.
Within each chunk, qualifying adjacent pairs are processed in descending
similarity order, with each token participating in at most one merge.
For a chunk of $n$ tokens, at most $\lfloor n/2\rfloor$ pairs
are accepted, so at least $\lceil n/2\rceil$ tokens remain.
Merging is performed once, without recursive or cross-chunk consolidation.

\paragraph{Sequence and position handling.}
In the Qwen integration, a common keep mask selects the updated
embeddings, token IDs, attention mask, modality labels, and multimodal
position IDs.
Text and other non-audio-visual tokens remain unchanged.
Retained tokens preserve their original semantic positions, with a merged
audio pair using the left token's position.
Physical cache slots are compacted separately, so sequence compaction does
not renumber the retained multimodal RoPE coordinates.
For MiniCPM-o-4.5, STC uses native modality bounds and treats each
64-token visual block as an $8\times8$ grid; such a block need not
correspond to a complete video frame. Its internal LLM handles positions,
so the Qwen-specific position-ID procedure is not assumed for this adapter.

\subsubsection{Semantic-Sensitivity Guided Precision Allocation}
\label{app:sspa_details}

\paragraph{Notation and page organization.}
We follow Sec.~\ref{sec:sspa}: $i$ indexes a retained token, $p$ a page,
$q$ a query anchor, and $t$ an inner allocation round.
Let $\mathcal P$ be the page set and $n_p$ the number of tokens in page $p$.
Each page contains at most 32 contiguous tokens and terminates at a modality
boundary; short pages are retained as indivisible allocation units.
The token indices belonging to page $p$ are denoted by $\mathcal I_p$.
For Algorithm~\ref{alg:sspa}, $\mathbf b=(b_p)_{p\in\mathcal P}$ denotes
the complete value-precision assignment, with $b_p\in\{2,4,16\}$.
Thus, $p$ denotes a page, whereas $\mathbf b$ denotes a complete candidate.
Within a refinement pass, the layout dependence of $\hat K$ and $C_a$
is implicit, as in the main text.
Head indices are omitted: attention scores are averaged across heads,
output energies are averaged across heads and channels, and page precision
assignments are shared across heads.

\paragraph{Query anchors and layer-group sharing.}
We use the final $N_q=\min(8,N)$ available prefill queries, where
$N=\sum_{p\in\mathcal P}n_p$ is the retained sequence length.
They are obtained from the corresponding input states using the allocation
layer's query projection and positional encoding.
Calibration attends to the complete retained history without an
anchor-specific prefill causal mask, serving as a proxy for decoding.
No external calibration data or future generated queries are used.

Layers are grouped in consecutive blocks of four.
The first layer uses its own $Q,K,V$ to compute a complete page assignment,
including all allocation rounds and layout refinements.
The remaining layers reuse the page tiers but independently quantize their
own KV with their own scales and zero points.
For Qwen2.5-Omni-3B, all decoder layers use SSPA.
For Qwen2.5-Omni-7B, layers 0 and 1 retain BF16 KV; grouping remains
0--3, 4--7, and so on, with layer 0 computing the assignment used by
layers 2 and 3.
Model-specific evaluated configurations are listed in
Appendix~\ref{app:main_config_registry}.

\paragraph{Hadamard preconditioning and quantization formats.}
We apply a normalized
Walsh--Hadamard transform along the attention feature dimension.
Queries and keys receive the same orthogonal transform after rotary
position encoding, preserving their dot products before quantization.
Values are also transformed, and the inverse transform is applied to
the readout before the attention output projection.
All calibration quantities $Q,K,V$ and their reconstructions are expressed
in this transformed space.

In SSPA-managed historical caches, keys use per-channel affine INT4,
calibrated over physical blocks of 32 tokens in the default Qwen profile.
INT2 values use per-token affine quantization in groups of 32 channels,
with FP16 scales and integer zero points.
INT4 values use per-token symmetric quantization with FP32 scales;
BF16 values retain their transformed representation.
The value tier applies to the whole page, whereas quantization parameters
are computed at the granularity of the corresponding format.

\paragraph{Token-level semantic signals.}
Using the reference attention $\mathcal A(K)$ from Sec.~\ref{sec:sspa},
token query relevance is
\begin{equation}
s_i^{\mathrm{attn}}
=\sum_{q=1}^{N_q}w_q[\mathcal A(K)]_{q,i},
\qquad
w_q=\frac{\gamma^{N_q-q}}{\sum_{j=1}^{N_q}\gamma^{N_q-j}}.
\label{eq:appendix_attention_reduction}
\end{equation}
Anchors follow input order, and $\gamma=0.8$ emphasizes later queries.
All selected anchors participate in this weighted average; no top-$k$
selection is applied to the anchors.

For alignment, audio and video tokens are independently assigned to
16 temporal chunks by relative sequence position after STC.
Chunks with the same index provide approximate cross-modal correspondence,
rather than exact timestamp synchronization.
Let $z_i$ be the normalized feature obtained by averaging token $i$'s
unquantized key features across heads, and let $\mathcal T(i)$ contain
tokens of the other modality in the matching chunk.
Then
\begin{equation}
s_i^{\mathrm{align}}
=\operatorname{TopMean}_{4}
\bigl(\{\cos(z_i,z_j):j\in\mathcal T(i)\}\bigr),
\label{eq:appendix_alignment}
\end{equation}
where $\operatorname{TopMean}_k$ averages the largest at most $k$ entries.
If $\mathcal T(i)$ is empty, we use cosine similarity to the other
modality's normalized global centroid.
If either audio or video is absent, alignment falls back to query relevance.

\paragraph{Page aggregation and semantic prior.}
Each signal is aggregated separately from tokens to pages and then
rank-normalized within audio or video:
\begin{equation}
s_p^{\xi}
=\operatorname{RankNorm}_{m(p)}\!\left(
\operatorname{TopMean}_{4}
\bigl(\{s_i^{\xi}:i\in\mathcal I_p\}\bigr)\right),
\qquad \xi\in\{\mathrm{attn},\mathrm{align}\}.
\label{eq:appendix_page_scores}
\end{equation}
Here, $m(p)$ is the modality of page $p$.
$\operatorname{RankNorm}_{m(p)}$ maps ranks among pages of that modality
to $[0,1]$, with tied scores receiving their average rank.
This prevents the raw scales of attention and alignment from dominating
their combination in Eq.~\ref{eq:sspa_semantic_prior}.
The default attention weight is $\omega=0.6$; text and other pages receive
a neutral semantic prior $s_p^{\mathrm{sem}}=0.5$.
The two top-4 operations have distinct roles: one selects cross-modal
similarities for each token, and the other aggregates token scores
within each page.

\paragraph{Attention-output distortion.}
\label{app:sspa_distortion}
For an output-space tensor $X$, let $X_q$ denote its component at anchor
$q$, and let $\|X_q\|_{\mathrm{av}}^2$ denote its mean squared energy.
$X$ can be the current residual $R_t$ or an upgrade correction $C_a$.
For $N_q>1$, define $\Delta_1X=0$ and
$\Delta_qX=X_q-X_{q-1}$ for $q>1$.
We use
\begin{equation}
\begin{aligned}
\ell_q(X)
&=(1-\lambda)\|X_q\|_{\mathrm{av}}^2
+\lambda\frac{N_q}{N_q-1}\|\Delta_qX\|_{\mathrm{av}}^2,\\
\mathcal L(X)
&=\operatorname{TopMean}_{k_\alpha}
\bigl(\{\ell_q(X)\}_{q=1}^{N_q}\bigr),
\qquad k_\alpha=\lceil(1-\alpha)N_q\rceil.
\end{aligned}
\label{eq:sspa_distortion}
\end{equation}
The top-tail mean implements the empirical upper-tail CVaR
objective~\citep{ying2022towards,kim2024risk}.
The factor $N_q/(N_q-1)$ compensates for the zero transition at the first
anchor. For $N_q=1$, we use $\mathcal L(X)=\|X_1\|_{\mathrm{av}}^2$.
Defaults $\alpha=0.75$ and $\lambda=0.5$ average the two largest losses
when $N_q=8$ and balance residual magnitude against anchor-to-anchor
variation. This is an attention-readout objective before the output
projection, not a final-logit or task-loss objective.
Anchor weighting in Eq.~\ref{eq:appendix_attention_reduction} is used
for relevance; the distortion objective uses the top-tail reduction above.

\paragraph{Action prior.}
For an action $a=(p,u\!\rightarrow\!v)$, we measure sensitivity by
$\mathcal L(C_a)$ and construct
\begin{equation}
P_a=\eta s_p^{\mathrm{sem}}+(1-\eta)
\operatorname{RankNorm}_{m(p)}\!\left(\mathcal L(C_a)\right).
\label{eq:appendix_action_prior}
\end{equation}
Correction sensitivities are normalized within each audio/video modality
for the corresponding upgrade type.
The defaults are $\eta=0.75$ and $\beta=0.25$ in
Eq.~\ref{eq:sspa_upgrade_priority}.
Thus, the prior multiplier lies in $[0.75,1.25]$ and adjusts the
distortion reduction per byte without admitting actions with $r_t(a)\leq0$.
The corrections $C_a$ and priors $P_a$ are computed once per refinement
pass; the gains $r_t(a)$ and priorities $J_t(a)$ are recomputed from
the current residual at every allocation round.

\paragraph{Precision budgets.}
Let $g$ index a layer group and $\bar b_g$ be its average value-bit ceiling.
For candidate $\mathbf b$, the nominal constraint is
\begin{equation}
\sum_{p\in\mathcal P} n_p b_p\leq N\bar b_g,
\qquad N=\sum_{p\in\mathcal P}n_p.
\label{eq:appendix_nominal_budget}
\end{equation}
Let $B_b$ denote the actual per-token value-format cost at tier $b$,
including its scales and zero points.
An upgrade $a=(p,u\!\rightarrow\!v)$ costs
$\Delta B_a=n_p(B_v-B_u)$.
The separate value-format byte constraint is
\begin{equation}
\begin{aligned}
\sum_{p\in\mathcal P}n_pB_{b_p}&\leq B_g^{\mathrm{budget}},\\
B_g^{\mathrm{budget}}
&=NB_2+\frac{\bar b_g-2}{14}N(B_{16}-B_2).
\end{aligned}
\label{eq:appendix_byte_budget}
\end{equation}
These expressions omit integer rounding of the implementation budgets.
The byte ceiling covers value payloads and their quantization metadata,
not key storage, page indices, Recent-FP, reserved capacity, or workspaces.
The bit budget is implemented in units of two additional bits per token:
INT2$\rightarrow$INT4 costs $n_p$ units and INT4$\rightarrow$BF16 costs
$6n_p$ units.

For normalized group depth $z_g\in[0,1]$, the ceiling follows
\begin{equation}
\bar b_g=b_{\mathrm s}+(b_{\mathrm d}-b_{\mathrm s})
\frac{\sigma(\kappa(z_g-1/2))-\sigma(-\kappa/2)}
{\sigma(\kappa/2)-\sigma(-\kappa/2)},
\label{eq:appendix_depth_budget}
\end{equation}
where $\sigma$ is the logistic sigmoid.
The default Qwen profile uses shallow/deep ceilings
$(b_{\mathrm s},b_{\mathrm d})=(3.6,3.0)$ and steepness $\kappa=8$.
The budget need not be exhausted.

For both byte and bit constraints, only the upgrade allowance above
the all-INT2 baseline is divided among modalities:
\begin{equation}
\rho_m=\frac{\nu_mN_m}{\sum_{m'}\nu_{m'}N_{m'}}.
\label{eq:appendix_modality_budget}
\end{equation}
Here, $N_m$ is the retained token count of modality $m$.
Default budget weights are $\nu_{\mathrm{video}}=0.6$,
$\nu_{\mathrm{audio}}=1.8$, and $\nu_{\mathrm{other}}=1.0$.
Integer remainders are distributed by largest fractional remainder;
unused modality allowances are not borrowed by other modalities.

\paragraph{Inner allocation rounds.}
\label{app:sspa_algorithm}
For fixed reconstructed keys $\hat K$, each allocation pass initializes
all page tiers to INT2 and sets $R_1=R_{\mathrm{init}}$ using
Eq.~\ref{eq:sspa_readout_actions}.
Algorithm~\ref{alg:sspa} runs at most $T_{\mathrm{alloc}}=4$ rounds.
At round $t$, each page contributes only its next one-tier action:
INT2$\rightarrow$INT4 or INT4$\rightarrow$BF16; BF16 pages have no action.
Actions with $r_t(a)\leq0$ are discarded.
Remaining actions are considered in descending $J_t(a)$ and accepted
only if their costs fit the remaining global, modality, and round budgets.
Infeasible actions are skipped so that less costly actions can still be selected.

The round allowance approximately divides the remaining upgrade budget by
the remaining number of rounds; the final round can use all remaining
allowance. Pages are indivisible, and no page advances twice in one round.
All actions in the selected batch $\mathcal U_t$ are scored against the
same starting residual $R_t$.
After applying the batch, we update
\begin{equation}
R_{t+1}=R_t-\sum_{a\in\mathcal U_t}C_a.
\label{eq:appendix_residual_update}
\end{equation}
Remaining actions are rescored using this updated residual.
An empty batch terminates the inner allocation.
Positive individual gains do not guarantee a decrease for the complete
batch because its corrections can interact; candidate evaluation below
therefore uses the complete reconstructed readout.

\begin{algorithm}[t]
\caption{Output-Sensitive Budgeted Precision Allocation}
\label{alg:sspa}
\small
\begin{algorithmic}[1]
\REQUIRE Historical $K,V$; query anchors $Q$; pages $\mathcal P$;
semantic priors $s_p^{\mathrm{sem}}$; precision budgets
\ENSURE Page precision assignment $\mathbf b^*=(b_p^*)_{p\in\mathcal P}$,
shared within the layer group
\STATE Set $b_p\leftarrow2$ for all pages; initialize candidate set $\mathcal S$
\STATE Evaluate $D(\mathbf b)$ under its own cache layout
(Eq.~\ref{eq:sspa_candidate_distortion}); add $(\mathbf b,D(\mathbf b))$ to $\mathcal S$
\FOR{$o=1,\ldots,T_{\mathrm{outer}}$}
  \STATE Reconstruct INT4 keys $\hat K$ under the layout induced by $\mathbf b$
  \STATE Compute $R_{\mathrm{init}}$, corrections $C_a$
  (Eq.~\ref{eq:sspa_readout_actions}), and priors $P_a$
  \STATE Set $b'_p\leftarrow2$ for all pages and $R_1\leftarrow R_{\mathrm{init}}$
  \FOR{$t=1,\ldots,T_{\mathrm{alloc}}$}
    \STATE Enumerate the next one-tier actions allowed by $\mathbf b'$;
    compute $r_t(a)$ and $J_t(a)$
    \STATE In descending $J_t(a)$, greedily form $\mathcal U_t$ from
    positive-gain actions satisfying global, modality, and round budgets
    \IF{$\mathcal U_t=\varnothing$}
      \STATE \textbf{break}
    \ENDIF
    \STATE Apply $\mathcal U_t$ to $\mathbf b'$;
    $R_{t+1}\leftarrow R_t-\sum_{a\in\mathcal U_t}C_a$
  \ENDFOR
  \STATE Set $\mathbf b\leftarrow\mathbf b'$; evaluate $D(\mathbf b)$ under its own layout
  \STATE Add $(\mathbf b,D(\mathbf b))$ to $\mathcal S$
\ENDFOR
\STATE Return the minimum-distortion assignment $\mathbf b^*$ in $\mathcal S$
(ties favor lower token-weighted value precision)
\end{algorithmic}
\end{algorithm}

\paragraph{Physical layout and outer refinement.}
Historical values are stably grouped into BF16, INT4, and INT2 pools.
Keys follow the same token permutation in a single compact INT4 arena,
where consecutive physical tokens form the key calibration blocks.
Logical token positions and their rotary coordinates remain unchanged.
Different page assignments can change the members of these blocks,
especially near short pages, and therefore change the reconstructed $\hat K$.

Each outer pass takes the current candidate $\mathbf b$, reconstructs
$\hat K$ from the original BF16 keys using that candidate's physical
blocks, and fixes these reconstructed keys during inner allocation.
It then restarts the value assignment at INT2, recomputes
$R_{\mathrm{init}}$, $C_a$, and $P_a$, and produces a new candidate.
Resetting the value tiers does not reset the fixed key reconstruction
used to score that pass.
The default $T_{\mathrm{outer}}=2$ permits two proposals in addition
to the initial all-INT2 candidate.

For each candidate $\mathbf b$, reconstruct its keys $\hat K$ under
its own layout and let $\hat V_{\mathbf b}$ contain its assigned value tiers.
Its final score is
\begin{equation}
D(\mathbf b)=\mathcal L\!\left(
\mathcal A(K)V-\mathcal A(\hat K)\hat V_{\mathbf b}\right).
\label{eq:sspa_candidate_distortion}
\end{equation}
This score uses mixed-precision values, unlike $R_{\mathrm{init}}$,
which always uses the all-INT2 baseline.
The candidate set $\mathcal S$ contains the initial assignment and every
proposal; the smallest $D(\mathbf b)$ is selected, with ties broken by
the smaller token-weighted average $N^{-1}\sum_p n_pb_p$.
Earlier proposals remain candidates even after another refinement pass.
If a reordering leaves key-block membership unchanged, calibration can be
reused and redundant refinement avoided.
Key reconstruction uses the runtime quantizer; value corrections are
simulated primarily in FP32, so stored-scale rounding and kernel
accumulation can introduce small differences from runtime execution.

\paragraph{Runtime and kernels.}
SSPA preserves the native FlashAttention-2 prefill output for the
STC-compressed input.
After each layer's prefill attention completes, its historical cache is
packed into the compact INT4 key arena and the three value pools.
Only the page assignment is shared across a layer group; cache values
and quantization parameters remain layer-specific.

During decoding, new queries, keys, and values undergo the corresponding
position handling and Hadamard transforms.
New KV states enter Recent-FP in BF16.
The mixed-precision Triton kernel reads the packed historical pools and
Recent-FP, dequantizes blocks on chip, and combines their local softmax
statistics into a single globally normalized readout.
The inverse value transform is applied before the output projection.
The full historical cache is not materialized in BF16 during decoding,
although prefill calibration and packing temporarily use floating-point KV.

Recent-FP is initially empty and has a default capacity of 128 tokens.
When full and another token must be appended, its aligned 128-token block
is transferred to history using INT4 keys and INT2 values.
This is blockwise flushing, not a sliding window that always retains
128 floating-point tokens.
Prompt-page assignments remain fixed during decoding, and transferred
generated tokens do not trigger semantic rescoring or reallocation.
When space is available, appending the current KV state is fused with
attention readout.
All scoring, allocation, and packing overheads are included in TTFT.

\begin{table}[t]
\centering
\small
\setlength{\tabcolsep}{5pt}
\renewcommand{\arraystretch}{1.08}
\caption{\textbf{Default R$^2$-Omni implementation parameters.}}
\label{tab:r2omni_hyperparameters}
\resizebox{\textwidth}{!}{%
\begin{tabular}{lll}
\toprule
\textbf{Component} & \textbf{Parameter} & \textbf{Default setting} \\
\midrule
Video STC
& Spatial / temporal similarity threshold
& $0.70$ / $0.70$ \\
Video STC
& Region size / prototypes / isolated retention
& $[2,128]$ / one per region / $100\%$ \\
Audio STC
& Chunk size / merge threshold / maximum merge ratio
& $50$ / $0.80$ / $0.50$ \\
\midrule
Semantic prior
& Query anchors / recency decay
& $8$ / $0.8$ \\
Semantic prior
& Temporal chunks / alignment top-$k$ / page top-$k$
& $16$ / $4$ / $4$ \\
Semantic prior
& Attention--alignment weight $\omega$
& $0.6$ \\
\midrule
Page layout
& Page size / layer-group size
& $32$ / $4$ \\
Key quantization
& Precision / scheme / calibration block
& INT4 / per-channel affine / $32$ tokens \\
Value quantization
& Tiers / INT2 group size
& BF16, INT4, INT2 / $32$ channels \\
\midrule
Distortion
& CVaR level $\alpha$ / transition weight $\lambda$
& $0.75$ / $0.5$ \\
Priority
& Semantic weight $\eta$ / prior strength $\beta$
& $0.75$ / $0.25$ \\
Allocation
& Outer refinements / allocation rounds
& $2$ / $4$ \\
Modality budget
& Video / audio / other weights
& $0.6$ / $1.8$ / $1.0$ \\
Depth schedule
& Shallow--deep ceiling / sigmoid steepness
& $3.6\!\rightarrow\!3.0$ bits / $8$ \\
Generated cache
& Recent-FP recompression chunk
& $128$ tokens \\
\bottomrule
\end{tabular}%
}
\end{table}

\subsection{Reproduction details of compared baselines}
\label{app:baseline_adapters}

KIVI and OTT do not natively support Qwen2.5-Omni or the internal Qwen3 model
used by MiniCPM-o.
We denote our model-side adaptations as KIVI-OM and OTT-OM.
The multimodal encoders, processor, and generation interface remain unchanged.
Only text self-attention and cache management are adapted.

\paragraph{Shared model interface.}
Prompt prefilling follows the original pipeline of Omni-LLMs with FlashAttention-2.
The adapter applies the model's native Q/K/V projections, Qwen2.5 multimodal
RoPE or Qwen3 QK normalization and RoPE, and then converts the completed cache
to the baseline's packed format.
During decoding, it maps Qwen GQA heads to the layout expected by the packed
operator and reconnects the resulting attention output to the original output
projection.
A thin cache wrapper exposes the packed state through the Transformers cache
interface.

For Qwen2.5-Omni-7B, queries are scaled by $1/\sqrt d$ before the FP16 packed
QK readout.
This is algebraically equivalent to scaling the completed logits and avoids
FP16 overflow without modifying the cache format or CUDA kernels.
Dense SDPA and FlashAttention-2 readouts are retained only as correctness
references; efficiency results use native packed decode.

\paragraph{KIVI-OM.}
KIVI-OM preserves KIVI's asymmetric quantization: keys are quantized
per-channel over token groups, while values are quantized per-token over
channel groups.
Packing and low-bit readout use KIVI's Triton quantizer and CUDA GEMV.
The adapter maintains the full-precision residual cache, appends newly
generated states, and flushes complete groups into the packed cache.

The reported Qwen2.5-Omni-3B experiments use the established
Qwen-compatible KIVI transition in which K and V share the same Recent-FP
boundary and aligned groups are quantized together.
The released KIVI state transition is also exposed as a separate audit mode.
We therefore report this baseline as KIVI-OM rather than claiming native
Qwen2.5-Omni support.

\paragraph{OTT-OM.}
OTT-OM directly invokes the released OTT cache update routine for residual
flushing, outlier scoring, traced-token updates, replacement, and K/V packing.
Its Triton packer and CUDA GEMV are imported from the upstream implementation
without modification.
The adapter only translates the Qwen cache to the tuple consumed by the
released update routine and maps Qwen GQA heads to the kernel layout.

Traced K/V states are restored in full precision during attention, while the
remaining cache is read from the packed representation.
Layer-wise tracing rules, residual size, outlier count, and auxiliary capacity
follow the evaluated OTT profile.
We report the resulting method as OTT-OM because the interface to Omni-LLMs,
multimodal positional encoding, and GQA integration are provided by our
adapter.

\subsection{Method Configurations at the Nominal \texorpdfstring{$5\times$}{5x} Budget}
\label{app:main_config_registry}

Triples follow VS (Video-SALMONN~2), LV (LongVALE), and OD (OmniDCBench) order.
$G$ is the quantization group size, $R$ the BF16 residual length, and
$P$ the set of protected BF16 layers.

\paragraph{R$^2$-Omni.}
Defaults follow Table~\ref{tab:r2omni_hyperparameters}.

$(V_s,V_d)$ denotes shallow/deep value payload ceilings; $G_K=G_V=G$.
\begin{itemize}\setlength{\itemsep}{0pt}
\item \textbf{Qwen3B:} $(V_s,V_d)=(3.6,3.0)$, $G=32$.
\item \textbf{Qwen7B:} $(V_s,V_d)=(3.6,3.0)$, $G=32$, $P=\{0,1\}$.
\item \textbf{MiniCPM:} ceilings $(2.1,2.0)$, $(3.0,2.4)$, $(2.2,2.0)$;
$G=(128,32,32)$; $P=\varnothing$ on VS and $\{0\}$ on LV/OD.
\end{itemize}

\paragraph{KIVI.}
All models use the official native packed adapter, INT2 K/V.
\begin{itemize}\setlength{\itemsep}{0pt}
\item \textbf{Qwen3B:} $R=1024$;
$G=64$ on VS and LV and 128 on OD.
\item \textbf{Qwen7B:}  $G=128$, $R=1024$, $P=\{0\}$.
\item \textbf{MiniCPM:} $G=128$, $R=384$.
\end{itemize}

\paragraph{OTT.}
All models use the official native packed adapter, INT2 K/V, $G=128$,
and sink capacity 32; $s$ denotes traced outliers per flush.
\begin{itemize}\setlength{\itemsep}{0pt}
\item \textbf{Qwen3B:} $R=(1280,1664,1408)$, $s=5$.
\item \textbf{Qwen7B:} $R=896$, $s=3$, $P=\{0\}$.
\item \textbf{MiniCPM:} $R=256$, $s=3$.
\end{itemize}

\paragraph{OScaR.}
All models use the official packed path, residual block size 128,
Hadamard rotation, key normalization, INT2 K/V, and $G=32$.

\paragraph{OmniSelect.}
AudioCLIP threshold is 2.0. 
Audio/video retention rates are equal. For all models, we retain 18\% on VS and LV, 12\% on OD.

\paragraph{OmniZip.}
Audio/video retention rates are equal. For all models, we retain 18\% on VS and LV, 12\% on OD.

\paragraph{SEATS.}
Audio/video retention rates are equal. For all models, we retain 18\% on VS and LV, 12\% on OD.

\subsection{R\texorpdfstring{$^2$}{2}-Omni Compression-Ratio Configurations}
\label{app:sweep_configs}
Let $\boldsymbol{\tau}=(\tau_s,\tau_t,\tau_a)$
denote spatial/frame-pair/audio thresholds.

\textbf{$3\times$:} $\boldsymbol{\tau}=(0.90,0.90,0.95)$;
ceilings $(6.1,4.8)$ on VS and $(6.5,5.2)$ on LV/OD.

\textbf{$4\times$:} $\boldsymbol{\tau}=(0.80,0.80,0.88)$;
ceilings $(4.7,3.8)$ on VS/LV and $(4.4,3.6)$ on OD.

\textbf{$6\times$:} $\boldsymbol{\tau}=(0.70,0.70,0.80)$;
ceilings $(3.9,3.2)$, $(3.7,3.0)$, and $(3.4,2.8)$ on VS/LV/OD.

\textbf{$7\times$:} $\boldsymbol{\tau}=(0.675,0.675,0.77)$;
ceilings $(3.2,2.4)$, $(2.8,2.2)$, and $(2.0,2.0)$ on VS/LV/OD.

\textbf{$8\times$:} $\boldsymbol{\tau}=(0.65,0.65,0.75)$;
ceilings $(3.2,2.4)$, $(2.8,2.2)$, and $(2.0,2.0)$ on VS/LV/OD.

\textbf{$9\times$:} $\boldsymbol{\tau}=(0.62,0.62,0.71)$;
ceilings $(3.2,2.4)$, $(2.8,2.2)$, and $(2.0,2.0)$ on VS/LV/OD.

\subsection{Pruning--Quantization Composition Configurations}
\label{app:composition_configs}

\paragraph{OmniZip + KIVI.}
OmniZip retains 65\% per modality, with $g=3$, contextual ratio 0.05,
and retention bounds 45\%--85\%. KIVI uses INT2 K/V,
$G=128$, and $R=(2944,3072,1152)$.

\paragraph{OmniSelect + OTT.}
OmniSelect retains 65\% per modality with AudioCLIP threshold 2.0.
OTT uses INT2 K/V, $G=128$, $R=(2944,3072,1408)$,
$s=5$, and sink capacity 32.

\paragraph{SEATS + OScaR.}
SEATS retains 70\% per modality with late-block selection disabled.
OScaR uses INT2 K/V, $G=32$, residual block size 128,
Hadamard rotation, and key normalization.

\section{Long-Form Generation Benchmarks}
\label{sec:benchmarks}

The experiments are conducted on the following long-form generation benchmarks.

\paragraph{video-SALMONN 2 Testset.}
The video-SALMONN2 testset~\citep{tang2025videosalmonn2} is a long-form generation benchmark designed to evaluate the completeness and factual accuracy of detailed video descriptions.
It contains 483 videos spanning 14 diverse domains, with durations ranging from 30 to 60 seconds and an average duration of approximately 51 seconds.
Each video is annotated with a set of fine-grained audio-visual events that serve as reference information for evaluating generated captions.
The official evaluation employs an LLM-based judge to identify missed, incorrect, and hallucinated events, from which the event error rate (EER) is computed.
We report the Correct rate and EER to jointly measure information preservation and generation accuracy.

\paragraph{LongVALE.}
LongVALE~\citep{geng2025longvale} is used as a long-form generation benchmark for temporally grounded descriptions of long videos.
The full dataset contains approximately 8.4K long videos with 105K omni-modal events, where each event is annotated with precise temporal boundaries and detailed relation-aware captions covering visual, audio, and speech information.
Its evaluation split consists of 1,171 videos containing 13,867 annotated omni-modal events, corresponding to 75.6 hours of video.
LongVALE supports three fine-grained tasks: omni-modal temporal video grounding, dense video captioning, and segment captioning.
Following the official Omni-DVC evaluation protocol, we use SODA-C to assess the overall quality of temporally grounded dense descriptions.

\paragraph{OmniDCBench.}
OmniDCBench~\citep{yao2026omnidcbench} is a human-annotated long-form generation benchmark that requires Omni-LLMs to generate continuous, fine-grained, and structured audio-visual narratives with explicit temporal boundaries.
It contains 1,122 high-resolution videos annotated by experts with cinematography knowledge, with an average annotation length of approximately 995 words per video.
The annotations follow a six-dimensional structural schema covering audio-visual events, background and environment, camera state, shot editing, dialogue, and acoustic cues.
This long-form generation benchmark jointly evaluates multi-scene temporal segmentation and detailed caption quality.
We report SodaM for the overall quality.

\paragraph{Generation Prompts.}
\label{app:generation_prompts}
We provide the complete generation prompts below.

\begin{benchmarkprompt}{Video-SALMONN 2: Generation Prompts}
Variant 1 (84 samples)
Please provide a thorough description of all the content in the video, including every detail. As you describe, ensure that you also cover as much information from the audio as possible, and be mindful of the synchronization between the audio and video as you do so.

Variant 2 (62 samples)
Provide a comprehensive description of all the content in the video, leaving out no details. Be sure to include as much of the audio information as possible, and ensure that your descriptions of the audio and video are closely aligned.

Variant 3 (59 samples)
Please describe all the information in the video without sparing every detail in it. As you describe, you should also describe as much of the information in the audio as possible, and pay attention to the synchronization between the audio and video descriptions.

Variant 4 (70 samples)
Thoroughly describe everything in the video, capturing every detail. Include as much information from the audio as possible, and ensure that the descriptions of both audio and video are well-coordinated.

Variant 5 (71 samples)
Offer a detailed description of the video, making sure to include every detail. Also, incorporate as much information from the audio as you can, and ensure that your descriptions of the audio and video are in sync.

Variant 6 (76 samples)
Describe every aspect of the video in full detail, covering all the information it contains. Additionally, include as much of the audio content as you can, and make sure your descriptions of the audio and video are synchronized.

Variant 7 (61 samples)
Give a detailed account of everything in the video, capturing all the specifics. While doing so, also include as much information from the audio as possible, ensuring that the descriptions of audio and video are well-synchronized.
\end{benchmarkprompt}

\medskip

\begin{benchmarkprompt}{LongVALE: User Prompt}
Could you please detail the events that took place during different time segments in the video? List the events in the format: From xx to xx, event1.
From xx to xx, event2.
... Use integer timestamps normalized to the range 00 to 99. Include relevant visual events, speech, sounds, and their relationships.
\end{benchmarkprompt}

\medskip

\begin{benchmarkprompt}{OmniDCBench: User Prompt}
First identify the meaningful scene changes throughout this entire {duration}-second video, then return only the final caption as valid JSON.
The output must contain 6 to 12 chronological JSON objects, begin at 00:00, and cover the video through {end}. The final timestamp must end at {end}. Use scene boundaries rather than fixed-duration slicing. Never generate a timestamp beyond the video.
Every object must contain exactly these string fields: timestamp, segment_detail_caption, video_background, acoustics_content, shooting_style, speech_content, camera_state. Use an empty string only when the video provides no evidence.
This is a syntax example for a different, hypothetical 10-second video; do not copy its timestamps or captions:
[
  {"timestamp":"00:00-00:04","segment_detail_caption":"First event.","video_background":"First setting.","acoustics_content":"First sounds.","shooting_style":"First shot style.","speech_content":"First dialogue or empty string.","camera_state":"First camera state."},
  {"timestamp":"00:04-00:10","segment_detail_caption":"Final event.","video_background":"Final setting.","acoustics_content":"Final sounds.","shooting_style":"Final shot style.","speech_content":"Final dialogue or empty string.","camera_state":"Final camera state."}
]
For the actual video, replace the example with 6 to 12 detailed objects covering 00:00-{end}. Output no Markdown or explanation. The first character must be [ and the last must be ]. Stop immediately after ].
\end{benchmarkprompt}

\section{Evaluated Omni-LLMs}
\label{app:model_architectures}

Efficient use of pretrained multimodal models supports broader
accessibility, including adaptation to scientific
domains~\citep{cao2025generalized}.
We briefly describe the two model architectures used in our experiments.

\paragraph{Qwen2.5-Omni.}
Qwen2.5-Omni~\citep{qwen25omnitechnicalreport} adopts a Thinker--Talker
architecture. The Thinker integrates visual and audio encoders with
a language backbone for multimodal understanding and text generation,
while the Talker generates speech from the Thinker's representations.
Audio and video tokens are temporally interleaved and processed with
time-aligned multimodal rotary position embeddings.
Our experiments use the Thinker, applying STC before language-model
prefill and SSPA to its historical KV cache.

\paragraph{MiniCPM-o-4.5.}
MiniCPM-o-4.5~\citep{cui2026minicpm} combines a SigLIP visual encoder,
a Whisper audio encoder, and a Qwen3-8B language backbone.
A visual resampler and an audio projector convert modality features
into embeddings for joint processing with text.
Dedicated speech decoders support audio generation.
R$^2$-Omni applies STC to the fused input embeddings and SSPA to the
language backbone's historical KV cache.

\section{Qualitative Case Studies}
\label{app:case_studies}

We provide qualitative examples to complement the quantitative evaluation and
illustrate how different compression strategies affect multimodal information
retention. Rather than comparing caption length, we focus on whether the
generated descriptions preserve reference-supported events and semantic details
throughout the input sequence.

\subsection{Long-Range Temporal Coverage}
\label{app:case_sold}

\begin{figure}[t]
    \centering
    \includegraphics[width=\linewidth]{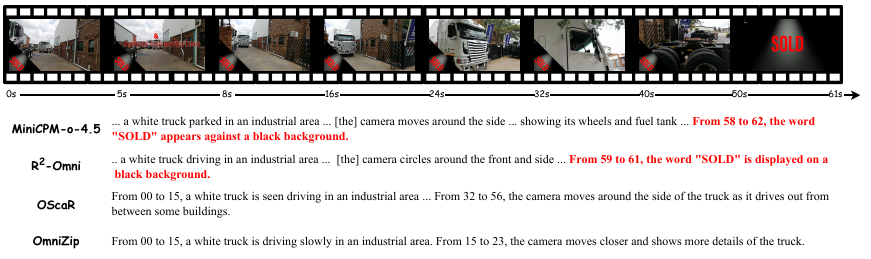}
    \caption{
    Qualitative comparison of long-range temporal coverage.
    R$^2$-Omni preserves more complete event coverage across the full clip,
    while OmniZip and OScaR terminate their semantic coverage earlier.
    The key evidence is highlighted in red.
    }
    \label{fig:case_sold}
\end{figure}

This example compares the temporal coverage preserved by different compression
strategies. As shown in Figure~\ref{fig:case_sold}, R$^2$-Omni maintains a
more complete description across the full video sequence, while OmniZip and
OScaR lose part of the temporally distributed information after compression.

The comparison shows that R$^2$-Omni is better able to preserve sparse events
appearing near the end of a long multimodal sequence, providing more complete
event coverage with a reduced memory footprint.

\subsection{Fine-Grained Detail Preservation}
\label{app:case_car}

\begin{figure}[t]
    \centering
    \includegraphics[width=\linewidth]{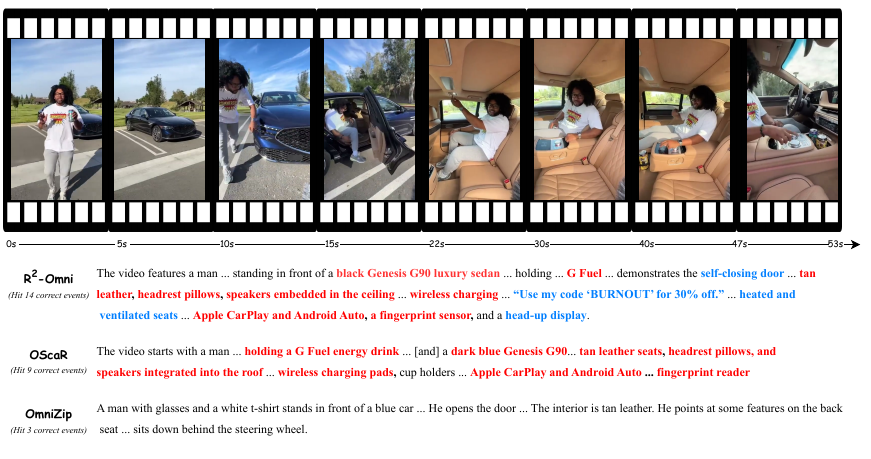}
    \caption{
    Qualitative comparison of fine-grained detail preservation.
    Red text highlights details preserved by both R$^2$-Omni and OScaR but
    omitted by OmniZip, while blue text highlights additional details preserved
    by R$^2$-Omni beyond OScaR.
    }
    \label{fig:case_car}
\end{figure}

This example compares how different compression methods preserve fine-grained
information in a detail-rich vehicle demonstration. As shown in
Figure~\ref{fig:case_car}, R$^2$-Omni retains substantially more specific
attributes and functional details than OmniZip, while also preserving
additional information beyond OScaR.

The comparison shows that R$^2$-Omni maintains a denser set of
reference-supported details under compression instead of reducing the video to
a coarse description of its overall progression. The official event judge
records 14 correct reference events for R$^2$-Omni, compared with 9 for OScaR
and 3 for OmniZip.

\subsection{Factuality under Compression}
\label{app:case_deer}

\begin{figure}[t]
    \centering
    \includegraphics[width=\linewidth]{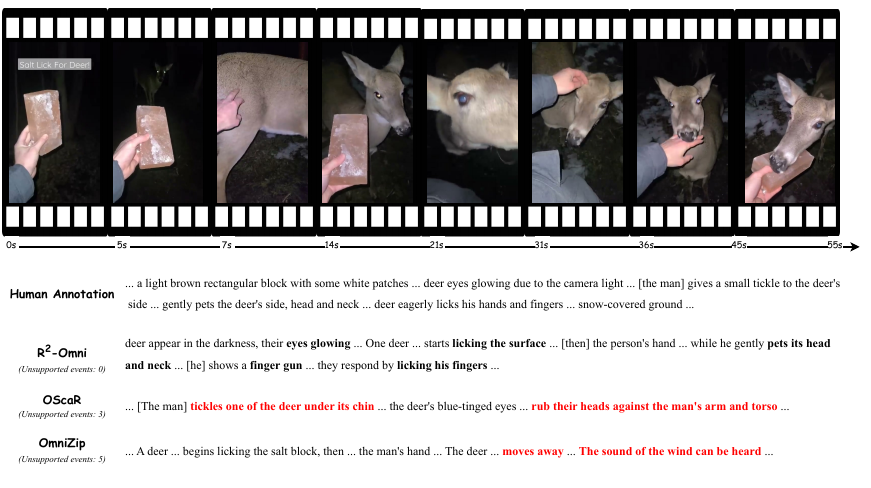}
    \caption{
    Qualitative comparison of factuality under compression.
    R$^2$-Omni matches BF16 with zero judge-flagged hallucinated events, while
    OmniZip and OScaR introduce additional unsupported content.
    Unsupported events are highlighted in red.
    }
    \label{fig:case_deer}
\end{figure}

This example examines factuality preservation under multimodal compression.
As shown in Figure~\ref{fig:case_deer}, R$^2$-Omni preserves the core
interaction while introducing no hallucinated events according to the official
event-level judge. The judge records zero hallucinations for both BF16 and
R$^2$-Omni, compared with five for OmniZip and four for OScaR.

The comparison suggests that removing redundant multimodal information can
preserve factual consistency and may reduce interference from irrelevant or
noisy representations during generation.

\subsection{Comparison Between STC and Token Pruning}
\label{app:case_stc}

\begin{figure}[t]
    \centering
    \includegraphics[width=\linewidth]{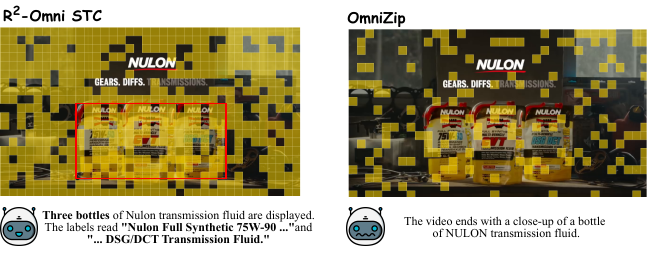}
    \caption{
    Comparison between STC and token pruning with the same final memory footprint.
    STC preserves substantially more original visual information after
    compression.
    }
    \label{fig:case_stc}
\end{figure}

This example compares STC with token pruning at the same final memory footprint.
As shown in Figure~\ref{fig:case_stc}, STC preserves substantially more of the
original visual information after compression, while token pruning leads to a
coarser representation and loses more fine-grained content.

This comparison shows that the advantage of STC does not come from using a
larger memory budget, but from preserving a denser set of useful visual
evidence with the same final memory footprint. This allows R$^2$-Omni to maintain
richer content with the same overall memory footprint.

\end{document}